%% file: main.tex
\documentclass[journal]{IEEEtran}
\ifCLASSINFOpdf
\else
\fi
\usepackage{times}
\usepackage{xcolor}
\usepackage{subcaption}
\usepackage{makecell}
\usepackage{latexsym}
\usepackage{hyperref}
\usepackage{booktabs}
\usepackage{siunitx}
\usepackage{tabularx}
\usepackage{array}
\usepackage{amsmath}
\usepackage{caption}
\usepackage{amsfonts}
\usepackage{adjustbox}
\usepackage{multirow}
\usepackage{tcolorbox}
\usepackage{enumitem}
\usepackage{xtab}
\usepackage{array}
\usepackage{longtable}
\usepackage[T1]{fontenc}

\usepackage[utf8]{inputenc}

\usepackage{microtype}

\usepackage{inconsolata}
\usepackage{algorithm}
\usepackage[noend]{algpseudocode} 
\usepackage[english]{babel}
\usepackage{amssymb}
\usepackage{tikz}

\usepackage{multirow}
\usepackage{array}

\usepackage{amsmath}
\usepackage{graphicx}
\usepackage{enumitem}
\usepackage{todonotes}
\usepackage{adjustbox}
\usepackage{booktabs}

\usepackage{soul}

\usepackage[acronym]{glossaries}
\glsdisablehyper 
\newcommand{\glslong}[1]{\glsdisp{#1}{\glsentrylong{#1}}}

\makeglossaries
\newacronym{ai}{AI}{Artificial Intelligence}
\newacronym{adas}{ADAS}{Advanced Driver-Assistance Systems}
\newacronym{ad}{AD}{Autonomous Driving}
\newacronym{odd}{ODD}{Operational Design Domain}
\newacronym{aeb}{AEB}{Automatic Emergency Braking}
\newacronym{eb}{EB}{Emergency Braking}
\newacronym{aes}{AES}{Automatic Emergency Steering}
\newacronym{lda}{LDA}{Lane Departure Assist}
\newacronym{acc}{ACC}{Adaptive Cruise Control}
\newacronym{tja}{TJA}{Traffic Jam Assist}
\newacronym{dnn}{DNN}{Deep Neural Network}
\newacronym{nn}{NN}{Neural Network}
\newacronym{rnn}{RNN}{Recurrent Neural Network}
\newacronym{gp}{GP}{Gaussian Process}
\newacronym{apf}{APF}{Artificial Potential Fields}
\newacronym{v2x}{V2X}{Vehicle-to-everything}
\newacronym{piml}{PIML}{Physics-informed machine learning}
\newacronym{ode}{ODE}{Ordinary Differential Equation}
\newacronym{sde}{SDE}{Stochastic Differential Equation}
\newacronym{pde}{PDE}{Partial Differential Equation}
\newacronym{cav}{CAV}{Connected and Autonomous Vehicle}
\newacronym{pinn}{PINN}{Physics-Informed Neural Network}
\newacronym{cbf}{CBF}{Control Barrier Function}
\newacronym{dcbf}{DCBF}{Discrete Control Barrier Function}
\newacronym{mpc}{MPC}{Model Predictive Control}
\newacronym{cf}{CF}{Car-Following}
\newacronym{p-ovm}{P-OVM}{Platoon Controlled OVM}
\newacronym{pcarnn}{PCARNN}{Physics-encoded Control-Affine Residual Neural Network}
\newacronym{dof}{DOF}{Degrees of Freedom}
\newacronym{pcarnn-dcbf}{PCARNN-DCBF}{Physics-encoded Control-Affine Residual Neural Network + Discrete Control Barrier Function}
\newacronym{com}{CoM}{Center of Mass}
\newacronym{ttc}{TTC}{Time-To-Collision}
\newacronym{zoh}{ZOH}{Zero-Order Hold}
\newacronym{rk4}{RK4}{Runge--Kutta}
\newacronym{qp}{QP}{Quadratic Programming}
\newacronym{cf1}{CF1}{Containment F1 Score}
\newacronym{fpr}{FPR}{False Positive Rate}
\newacronym{mcd}{MCD}{Median Containment Distance}
\newacronym{mcd+}{MCD$^+$}{Median Containment Distance with Signed Preference}
\newacronym{carla}{CARLA}{Car Learning to Act}
\newacronym{mrm}{MRM}{Minimum Risk Maneuver}
\newacronym{hil}{HIL}{Hardware-in-the-Loop}
\newacronym{tim}{TIM}{Traveler Information Messages}
\newacronym{brt}{BRT}{Backward Reachable Tube}

\newcommand{\pos}{\mathbf{p}}   
\newcommand{\sdf}{d}            
\newcommand{\barrier}{h}        

\input{graph_preamble}

\begin{document}
%
\title{PCARNN–DCBF: Minimal-Intervention Geofence Enforcement for Ground Vehicles}
%
%
%

\author{Yinan Yu 
        and Samuel Scheidegger
\thanks{Yinan Yu is with the Department
of Computer Science and Engineering, Chalmers University of Technology,
Sweden and Asymptotic AI, Sweden. E-mail: yinan@chalmers.se.}
\thanks{Samuel Scheidegger is with Asymptotic AI, Sweden.}}

\markboth{Journal of \LaTeX\ Class Files,~Vol.~xx, No.~x, November~2025}%
{Yu \MakeLowercase{\textit{et al.}}: PCARNN-DCBF}
%



\maketitle

\begin{abstract}
Runtime geofencing for ground vehicles is rapidly emerging as a critical technology for enforcing Operational Design Domains (ODDs). However, existing solutions struggle to reconcile high-fidelity learning with the structural requirements of verifiable control. We address this by introducing PCARNN-DCBF, a novel pipeline integrating a Physics-encoded Control-Affine Residual Neural Network with a preview-based Discrete Control Barrier Function. Unlike generic learned models, PCARNN explicitly preserves the control-affine structure of vehicle dynamics, ensuring the linearity required for reliable optimization. This enables the DCBF to enforce polygonal keep-in constraints via a real-time Quadratic Program (QP) that handles high relative degree and mitigates actuator saturation. Experiments in CARLA across electric and combustion platforms demonstrate that this structure-preserving approach significantly outperforms analytical and unstructured neural baselines.
\end{abstract}


\glsresetall

%
\IEEEpeerreviewmaketitle

\section{Introduction}

\IEEEPARstart{G}{eofencing} is a technique that uses virtual geographic boundaries to regulate vehicle operation within designated areas. It is the specification and runtime enforcement of spatial policies by predicting impending non-compliance and applying necessary interventions to stay compliant.
It shares similarities with other AD/ADAS functions, such as lane keeping, collision avoidance, and road-edge protection, but its purpose is distinct: it governs \emph{where} the vehicle is allowed to operate at all, not only \emph{how} it should follow lanes or avoid objects. In practice, geofencing serves as a policy and safety envelope that complements perception and planning by continuously assessing proximity to a prescribed region and applying the minimal correction required to avoid a boundary breach \cite{balachandran2018geofence}.

\paragraph{Representations}
A \emph{geofence} is a geometric region defined in a global or local coordinate frame. Two fundamental modes are \emph{keep-in} (remain inside the region) and \emph{keep-out} (avoid entering the region) \cite{cho2018assess}. A geofence are commonly expressed as a polygon, a corridor (the set between two boundaries), a level-set of a distance function, or a union of such sets. 
\begin{itemize}
\item Polygon: This is the most common representation for static, hard boundaries. Research in both ground and aerial vehicles frequently defines geofences using polygonal horizontal boundaries \cite{stevens2020generating}. While simple, this method must account for complex geometries; algorithms have been developed to handle "concave geozones" \cite{thomas2024geofencing}, as well as "multi-island" and "multi-hole" polygons \cite{hong2022new}. In their simplest form, enforcement is achieved via a Point-in-Polygon check \cite{thoren2024model}, though this is insufficient for predictive intervention. 
\item Corridor: This representation is intrinsically linked to motion planning. The literature on \gls{mpc} for autonomous driving, for example, explicitly uses this concept. An integrated approach may feature a "corridor path planner" that defines the "constraints on vehicle position" \cite{plessen2017spatial}. This corridor is then fed into the \gls{mpc} controller, which optimizes the vehicle's trajectory within those constraints. Here, the geofence is not a hard boundary but the dynamic input to an optimization-based controller.
\item Level-set: This is a more formal and flexible, though computationally intensive, approach. In robotics practice, such implicit representations are commonly instantiated as Signed Distance Fields (SDFs), which encode obstacle geometry with distances and gradients that integrate naturally with control, trajectory optimization, and learning methods \cite{bukhari2025differentiable,ortiz2022isdf,li2024configuration}. A level-set method defines the geofence boundary as the zero-value contour of a higher-dimensional implicit function \cite{hong2022new}. In robotics, this is used for path planning by defining an anisotropic cost function (e.g. risk) and solving for the optimal, minimum-risk path using methods that solve the Eikonal or Hamilton-Jacobi-Bellman equations \cite{wang2020robot}. Its primary advantage is in representing complex, dynamic boundaries, as the boundary can be evolved over time by updating the level-set function \cite{otte2015survey}.

\end{itemize}

\paragraph{Functional taxonomy}
The functional taxonomy of geofencing characterizes geofences along three independent axes: static versus dynamic, hard versus soft, and passive versus active, each describing a different operational property of how the geofence is defined, interpreted, and enforced.
\begin{itemize}
\item \emph{Static} / \emph{dynamic}: Static fences are fixed regions encoded in digital maps. This is the baseline assumption for most SAE Level 4 (L4) \gls{odd} deployments 17 and for safety-critical test tracks.4 Dynamic fences are temporary, context-dependent areas. The literature strongly associates this capability with \gls{v2x} and Cooperative Intelligent Transport Systems (C-ITS).18 Research has demonstrated \gls{v2x}-aided autonomous driving systems that use \gls{tim} to "detou[r] around a construction site ahead", a successful real-world execution of a dynamic geofence.
\item \emph{Hard} / \emph{soft}: Hard fences define strict invariance conditions that the vehicle must never violate. Soft fences introduce buffer zones or graded boundaries in which the system can modulate its response based on estimated risk or uncertainty. Soft implementations are often used when small deviations are tolerable or when measurement uncertainty warrants conservative, layered responses.
\item \emph{Passive} / \emph{active}: Passive fences issue warnings or advisories but leave the final action to the driver or upstream planner. Active fences initiate automatic intervention, such as braking, steering, or a coordinated maneuver, when a violation is predicted. Active enforcement requires actuator authority to reside on board the vehicle; external services may transmit geofencing constraints or stop requests via \gls{v2x}, but a safety-rated on-board controller remains the sole command owner for brake and steering execution.
\item \emph{Ground} / \emph{aerial} / \emph{maritime}:
For ground vehicles, geofencing typically acts in the planar $x$–$y$ space with heading and speed dynamics considered when predicting boundary crossings and planning evasive actions.
For aerial systems, geofences are 3D volumes in $(x,y,z)$ with altitude floors/ceilings and climb/sink rate constraints \cite{kim2022airspace, thomas2024geofencing}.
For maritime and underwater vehicles, they are surface regions in $(x,y)$ with shoreline and traffic-separation buffers or 3D $(x,y,z)$ volumes with depth limits and bathymetry constraints. In all cases, attitude/turn-rate and speed dynamics inform prediction and the choice of feasible interventions.
\end{itemize}

\paragraph{Violations}
Geofence violations may originate from different classes of events: (i) \emph{human factors}: driver distraction, drowsiness, or misjudgment; (ii) \emph{autonomy faults}: perception dropouts, localization drift, stale or mismatched maps, planner failures, or degraded actuation;
(iii) \emph{security events}: spoofed commands, GNSS manipulation, or malicious path injection that would drive the vehicle out of bounds, as exemplified in the broader automotive security literature \cite{checkoway2011comprehensive, gupta2023aninvestigation}. In all cases, geofencing aims to detect impending boundary violations early and apply the smallest safe correction.

\paragraph{Actuation choices}
When a geofence violation is predicted, \gls{aeb} and \gls{aes} are both valid responses, but they address different geometries and margins. \gls{aeb} is appropriate when the residual buffer allows the vehicle to stop within tire–road limits, or when lateral space is scarce or uncertain. \gls{aes} is preferable when there is sufficient lateral clearance to redirect the vehicle while staying within stability envelopes (yaw rate, sideslip, rollover) and without creating new hazards. In practice, the controller may need to choose between braking, steering, or a coordinated combination (e.g. brake-to-stabilize, then steer) based on real-time feasibility under friction limits and surrounding traffic conditions.

The context also matters. If the vehicle experiences a critical system failure, continuing operation is not an option and a controlled stop via \gls{aeb} is the appropriate response.
This type of brake-to-stop fallback is similar in spirit to the \glspl{mrm} discussed in automotive safety assessments built around standards such as ISO~26262 and ISO/SAE~21434, where automation transitions the vehicle toward a reduced-risk stopped condition when it can no longer guarantee safe operation (e.g. due to ODD exit or a critical failure)~\cite{volkswagen2023,UNECE2023,Balakrishnan2022}. 
In our setting, we focus on this control-level conservative fail-safe stop rather than a full, standards-defined \gls{mrm}.
By contrast, for scenarios such as autonomous testing or routine operation within a bounded service area, path correction through \gls{aes} may be preferable to preserve continuity of motion. Across all cases, a guiding principle is \emph{minimal intervention}: apply the smallest deviation necessary to guarantee compliance with the geofence while avoiding disruptive or unsafe maneuvers. Actuator command ownership must be unambiguous. Two deployment patterns cover most cases. (a) \emph{Vehicle-owned actuation}: the geofencing function runs on board or issues intervention requests (e.g. stop, lateral redirect with bounds) to the vehicle motion controller, which converts them into brake and steering commands under stability and friction limits. (b) \emph{Service-assisted geofencing}: an off-board service computes constraints or issues a stop request, but an on-board safety controller is the only entity permitted to actuate; external services never bypass this controller. In both patterns, actuator commands flow only through the on-board command owner; external inputs are advisory or supervisory and are validated before execution.
\paragraph{Architectural view}
A geofencing function can be understood in two stages: a \emph{predictive layer} that estimates whether the current state and control will lead to a boundary breach, and a \emph{control layer} that decides how to intervene, through braking, steering, or a combination, while respecting vehicle dynamics and safety constraints. Detailed design options for both stages are discussed in the background section.
Within this two-stage view, only the control layer holds actuator authority. The predictive layer may request an intervention (brake, steer, or coordinated) with target setpoints or bounds, but it does not issue torque or steering commands directly. If a geofencing service is off board, it interfaces at the request level; final actuation remains on board.
\paragraph{Applications}
Geofencing is directly connected to the \gls{odd}: it enforces the spatial component of the \gls{odd} at runtime~\cite{thoren2024model}. Beyond commercial deployments that restrict automated operation to validated service areas, several concrete scenarios motivate an explicit geofencing layer:
\begin{itemize}
  \item \textbf{Proving grounds and test tracks.} Keep a test vehicle within a validated area (e.g. a proving ground), stopping or steering when simulations or experiments push the vehicle toward an unsafe perimeter.
  \item \textbf{Geofenced automated services.} Low- to medium-speed automated shuttles, valet parking, campus logistics, mines, ports, and warehouses where the site boundary defines where autonomy is permitted.
  \item \textbf{Temporary or dynamic zones.} Road works, special events, school zones, and emergency scenes where the allowable region changes over time or moves with a convoy or escort.
  \item \textbf{Infrastructure safety margins.} Bridges, waterfronts, and drop-offs where boundary violations have high consequence and must trigger decisive action.
\end{itemize}

\paragraph{Relation to adjacent \gls{ad}/\gls{adas} topics}
Geofencing intersects but is not identical to lane departure prevention, collision avoidance, and road-edge detection. Lane departure focuses on markings and rules; geofencing governs arbitrary polygons and corridors. Collision avoidance manages moving/static objects; geofencing treats the workspace boundary as the constraint. Road-edge protection uses perception of drivable limits; geofencing can be defined via localization even without visual cues. At the supervisory level, geofencing aligns with \gls{odd} compliance, formal safety monitoring, and security measures that reject commands leading outside the allowed set.

In this paper, we consider ground vehicles with static, hard keep-in geofences given as polygons, and encode safety via a signed-distance level-set barrier $h(x)=d(p)$ on the vehicle position.
We focus primarily on braking and steering as the actuation strategy.

Our primary contribution is a novel geofencing pipeline, \gls{pcarnn-dcbf}, which integrates a structured, physics-encoded dynamics model with a predictive barrier-function controller for reliable, optimization-based enforcement of polygonal keep-in constraints under complex vehicle dynamics. Specifically, we introduce:
\begin{itemize}
\item \glslong{pcarnn}, a hybrid model that learns corrections for both the drift and control-effectiveness terms of the vehicle dynamics. This architecture strictly preserves the control-affine structure essential for tractable, optimization-based control.

\item A tractable, preview-based \glslong{dcbf} controller that \emph{builds on} sampled-data and zero-order \glspl{cbf} for safety-critical control~\cite{agrawal_discrete_2017,taylor_safety_2022,breeden_control_2022,tan_zero-order_2025} in the polygonal geofencing setting. It enforces a terminal barrier constraint on a numerically integrated zero-order-hold map, and combines a minimal-deviation \gls{qp} objective, and saturation-aware bound--secant linearization near actuator limits to achieve real-time geofence enforcement with minimal intervention.

\item We introduce a set of specialized metrics (\gls{cf1}, \gls{fpr}, \gls{mcd+}) to address the key requirements of accurate intervention, successful containment, and minimal correction. Using this framework, we conduct a comprehensive evaluation in the \gls{carla} simulator, where we benchmark the \gls{pcarnn-dcbf} pipeline against analytical, Neural \gls{ode}, and standard physics-encoded residual models. We analyze performance by examining control linearity and robustness across different vehicle types (electric vs. combustion) and challenging driving regimes.
\end{itemize}
The paper is structured as follows: Section \ref{sec:background} reviews the problem setting, vehicle dynamics models, and background on control barrier functions. Section \ref{sec:related} surveys related work in discrete-time \glspl{cbf} and learned dynamics. Section Section \ref{sec:method} details our proposed method, including the \gls{pcarnn} dynamics model and the preview-based \gls{dcbf} controller. Section \ref{sec:experiments} presents the experimental setup, benchmark comparisons, and a detailed analysis of the results.

\section{Background}
\label{sec:background}
\subsection{Problem setting and preliminaries}

We model geofencing as the runtime enforcement of a spatial constraint, defined by a region $S(t)\subseteq\mathbb{R}^2$ in the world frame. The set $S(t)$ denotes the admissible operating area at time $t$. For \emph{static} geofences, $S(t)$ reduces to a fixed region $S \subseteq \mathbb{R}^2$, such as a polygon extracted from a digital map, and the vehicle must remain inside $S$ for all time.

Let the vehicle position in the world frame be $\mathbf{p}=[p_{x},p_{y}]^\top \in \mathbb{R}^2$, where $p_x$ and $p_y$ denote the Cartesian coordinates in the world frame. Under \emph{keep-in} policies, admissible states satisfy $\mathbf{p}\in S$.

The full vehicle state in the world frame is defined as
\begin{equation}
  \label{eqa:full_state}
  x = [p_{x}, p_{y}, \psi, v_{x}, v_{y}, \omega, \delta]^\top \in \mathbb{R}^7
\end{equation}
where $\psi \in [-\pi,\pi]$ is the yaw angle, $v_x$ and $v_y$ are the longitudinal and lateral velocities in the body frame, $\omega$ is the yaw rate, and $\delta$ is the front-wheel steering angle.
This stacks the vehicle pose and dynamic bicycle states in a form consistent with standard dynamic single-track models~\cite{rajamani2012vehicle,polack2017,patil2017generic}.

We define a geometric signed distance $\sdf(p)$ on position and a state barrier $h(x)$ on the full state. Specifically,
$\sdf:\mathbb{R}^2\!\to\mathbb{R}$ gives the signed distance from the position $p=[p_x,p_y]^\top$ to the geofence boundary. 
Its zero level set
$\{\,p \in \mathbb{R}^2 : \sdf(p) = 0\,\}$ coincides with the boundary, with $\sdf(p) > 0$ for points
inside the admissible region $S$ and $\sdf(p) < 0$ for points outside.
The barrier on the full world-frame state $x$ is the
composition
\[
  h(x) := \sdf(p),\qquad p=\begin{bmatrix}p_x\\ p_y\end{bmatrix}.
\]
Constructing control barrier functions directly from signed distance fields has been proposed as a practical strategy
for collision avoidance, since the signed distance provides a scalar safety margin and its gradient supplies a natural
avoidance direction that can be embedded in control synthesis~\cite{wu2025optimization,koptev2023neural}.
For simple polygonal fences, $\sdf(p)$ is the minimum Euclidean distance to the polygon edges, with the sign determined by
an even–odd (ray-crossing) inside test (orientation-independent)~\cite{hormann2001point}.

Since $\sdf$ is positive inside the safe zone, $\nabla \sdf$ points inward (the gradient is undefined only at measure-zero sets such as vertices). We define the outward and inward unit normals as $\mathbf n_{\text{out}}=-\nabla\sdf/\|\nabla\sdf\|$ and $\mathbf n_{\text{in}}=+\nabla\sdf/\|\nabla\sdf\|$, respectively.

Symbols and definitions used in this paper can be found in Table~\ref{tab:symbols} for convenience.
\begin{table}[!ht]
\caption{Symbols and definitions}
\label{tab:symbols}
\begin{tabular}{p{2.3cm} p{5.7cm}} 
\toprule
  \multicolumn{2}{l}{\textbf{States and Parameters}} \\
World frame & A global coordinate system fixed to the environment, used to express the vehicle’s absolute position and orientation. \\
Body frame & A local coordinate system attached to the vehicle, with the $x$-axis pointing forward (longitudinal) and $y$-axis pointing to the left (lateral), used to express velocities and dynamic states. \\
  $p_{x}, p_{y}$ & Vehicle coordinates in the world frame.\\
  $\mathbf{p}\in \mathbb{R}^{2}$ & Location vector of the vehicle $\pos=[p_x,p_y]^\top$ in the world frame.\\
  $m$ & Vehicle mass. \\
  $v_{x}$ & Longitudinal velocity of the vehicle in its body frame (m/s). \\

  $v_y$ & Lateral velocity of the vehicle in its body frame (m/s). \\
  $\omega$ & Yaw rate of the vehicle (rad/s), representing rotation around the vertical axis.\\
  $\delta$ & Front wheel steering angle (rad).\\
  $\psi \in [-\pi, \pi]$ & Vehicle’s yaw angle in the world frame. \\
    $R(\psi)$ & Rotation matrix from body to world frame: $\begin{bmatrix} \cos{\psi} & -\sin{\psi} \\ \sin{\psi} & \cos{\psi} \end{bmatrix}$. The transpose of this matrix is the rotation matrix from world to body frame.\\
  ${x} \in \mathbb{R}^7$ & Vehicle’s full state in the world frame: $x=[p_{x}, p_{y}, \psi, v_{x}, v_{y}, \omega, \delta]^{\top}$. \\
  $\dot{x}$ & Derivative of the world frame state. \\
  $\sdf(\pos)$ & Signed distance from $\pos$ to a simple polygon geofence: the minimum Euclidean edge distance with sign from an orientation-independent even–odd (ray-crossing) inside test \cite{hormann2001point}; positive inside, negative outside. \\
  $\mathbf n_{\text{out}}(\pos)$ & Outward unit normal at $\pos$: $\mathbf n_{\text{out}}=-\nabla\sdf(\pos)/\|\nabla\sdf(\pos)\|$. \\
  $\barrier(x)$ & State barrier defined as $\barrier(x)=\sdf(\pos)$.\\
  $\mathbf{x}$ & Vehicle’s dynamic state in its body frame. The exact composition depends on the model. This composition could be, for example, $[v_x, v_y, \omega, \delta]^{\top}$. \\
  $\dot{\mathbf{x}}$ & Derivative of the body-frame state $\mathbf{x}$.\\
  ${r}$ & Parameters of physics models (e.g. for the bicycle model, $r =(\texttt{mass}, I_{z}, C_{f}, ...)$). \\
  $\alpha_f, \alpha_r$ & Slip angles at the front and rear wheels. \\
  $F_{yf}, F_{yr}$ & Lateral tire forces at the front and rear wheels. \\
  \midrule
\multicolumn{2}{l}{\textbf{Control Model}} \\
$\mathbf{u} \in \mathbb{R}^2$ & Control input $\mathbf{u}=\begin{bmatrix}\dot{\delta} \\ F_x\end{bmatrix}$, where $\dot{\delta}$ is the commanded steering rate (rad/s) and $F_x$ is the applied longitudinal force (N). \\

$f(\mathbf{x}, r)$ or $f(\mathbf{x})$ & Drift vector field: the natural evolution of the system with parameters $r$ when no control input is applied ($\mathbf{u}=0$). \\
$g(\mathbf{x}, r)$ or $g(\mathbf{x})$ & Control effectiveness matrix: maps each control input in $\mathbf{u}$ to its influence on the state dynamics.\\
Control-affine model & A general nonlinear system structure where dynamics separate into a {drift vector field} and a {control effectiveness matrix} multiplying the control input $\mathbf{u}$: $\dot{x} = f(\cdot) + g(\cdot)\mathbf{u}$. \\
  \\
  \midrule
\multicolumn{2}{l}{\textbf{Neural Network}} \\
$\theta$ & Learnable weights and biases of the neural network. \\
$NN_\theta(\cdot)$ & Neural network parameterized by $\theta$. Its specific inputs and outputs differ by architecture. \\
$\dot{\mathbf{x}}_{\text{pred}}$ & Predicted state derivative. \\
$\dot{\mathbf{x}}_{\text{gt}} \in \mathbb{R}^4$ & Ground truth state derivatives from real-world or simulated driving scenarios. \\
$\lambda_{\text{data}}$ & Weight of the data loss term (hyperparameter). \\
$\lambda_{\text{phys}}$ & Weight of the physics loss or regularization term (hyperparameter). \\
\bottomrule
\end{tabular}
  \end{table}

\subsection{Vehicle dynamics}

\subsubsection{Dynamic bicycle model}
\label{sec:bicycle}
The motion of a ground vehicle can be described by dynamic equations that relate the vehicle’s state to its control inputs, capturing how velocities and orientation evolve under steering and longitudinal actuation. In this work, we model the body-frame dynamics; the global pose kinematics (position and heading) are handled separately.

We adopt the \emph{dynamic bicycle model} as our representation of the vehicle dynamics. The bicycle model is a widely used reduced-order single-track abstraction that aggregates the left and right wheels of each axle into a single equivalent wheel, while preserving the essential lateral and longitudinal behavior relevant for control design~\cite{rajamani2012vehicle,polack2017,patil2017generic}. This model provides a tractable yet expressive approximation of the vehicle’s lateral–longitudinal dynamics for control and safety-critical applications.
However, this simplified single-track representation remains an approximation. For low- to moderate-lateral-acceleration regimes, dynamic bicycle models have been shown to capture the dominant coupling between longitudinal, lateral, and yaw dynamics with sufficient accuracy for control-oriented modeling and prediction~\cite{rajamani2012vehicle,polack2017,patil2017generic}. Beyond these nominal regimes, it neglects effects such as load transfer, additional body degrees of freedom, and coupled longitudinal–lateral tire dynamics that become important at higher speeds and in more aggressive transients~\cite{polack2017,patil2017generic}. Comparative studies indicate that low-order kinematic or single-track models perform well mainly under such nominal conditions with limited lateral acceleration, whereas more detailed multi-DoF models that include, e.g., combined longitudinal–lateral tire effects and load transfer are preferable as lateral acceleration approaches the vehicle’s handling limits or during rapid transient maneuvers~\cite{polack2017,patil2017generic}. Recent residual-learning work further shows that simple 3-DoF physics-based models tend to drift in long-horizon state prediction, and that learning residual corrections on top of such a base model can substantially reduce prediction error~\cite{miao2025residual}. Taken together, these observations motivate our choice to retain a low-order dynamic bicycle model for its tractability and analytic structure, while augmenting it with learned residual terms rather than replacing it with a high-dimensional physics model.

\input{piml_table}

The resulting bicycle model captures the longitudinal, lateral, and yaw motions of the vehicle. We adopt the standard approximation that lateral tire forces depend on slip angle but are independent of the commanded longitudinal force $F_x$. Under this assumption, the dynamics can be expressed in control-affine form as
\begin{equation}
  \label{eqa:control-affine}
  \dot{\mathbf{x}} = f(\mathbf{x}) + g(\mathbf{x})\mathbf{u},
\end{equation}
where $\mathbf{x} = [v_x, v_y, \omega, \delta]^\top$ is the dynamic state in the body frame and $\mathbf{u} = [\dot{\delta}, F_x]^\top$ is the control input consisting of steering rate and longitudinal force. 
We model the steering angle $\delta$ as a state, which makes the steering rate $\dot{\delta}$ the control input. This is a standard technique in vehicle handling and optimal-control formulations~\cite{HENDRIKX01121996,Massaro03072021}, as it provides a simple way to capture steering-actuator dynamics and enforce steering-rate limits. In our setting, this choice also preserves the control-affine structure of the dynamics, which we will exploit in the controller design of Sec.~\ref{subsec:method-controller}.

Further, we define the slip angles and lateral tire forces for the bicycle model.
The front and rear slip angles $\alpha_f, \alpha_r$ are given by
\[
\alpha_f = \operatorname{atan2}\!\big(v_y + \ell_f \omega,\ v_{x,\text{safe}}\big) - \delta,
~
\alpha_r = \operatorname{atan2}\!\big(v_y - \ell_r \omega,\ v_{x,\text{safe}}\big),
\]
where $\ell_f,\ell_r$ are the distances from the center of gravity to the front and rear axles, 
and $v_{x,\text{safe}}$ is a safeguarded longitudinal velocity defined as $v_{x,\text{safe}}=\mathrm{sign}(v_x)\,\max(|v_x|,\varepsilon)$ with $\varepsilon>0$ and $\mathrm{sign}(0)=+1$.
We compute lateral forces using the Pacejka Magic Formula (lateral-only) with the convention that positive slip yields negative lateral force~\cite{pacejka1992magic} (parameters $\mu$ friction, $C$ shape, $E$ curvature):
\begin{equation}
\label{eq:MF}
\small{F_y(\alpha) = -D\,\sin\!\Big(C\,\arctan\!\big(B\alpha - E\,(B\alpha - \arctan(B\alpha))\big)\Big)}
\end{equation}
Consistent with our implementation, we set the per-axle peak factor to
\[
D=\mu F_z,\qquad F_z=\tfrac{m g}{2}\ \text{(where $g$ is gravitational acceleration)},
\]
and derive front/rear stiffness factors from identified cornering stiffnesses $(C_f,C_r;\ \text{front/rear cornering stiffness, N/rad})$:
\begin{equation}
\label{eq:BfromCf}
B_f=\frac{C_f}{C\,D},\qquad B_r=\frac{C_r}{C\,D}.
\end{equation}

The natural evolution of the dynamics without control (i.e., the drift vector field) is
\begin{equation}
  \label{eqa:f}
f(\mathbf{x}) =
\begin{bmatrix}
-\tfrac{F_{yf}\sin\delta}{m} + v_y \omega \\
\tfrac{F_{yr} + F_{yf}\cos\delta}{m} - v_x \omega \\
\tfrac{\ell_f F_{yf}\cos\delta - \ell_r F_{yr}}{I_z} \\
0
\end{bmatrix},
\end{equation}
where $m$ is the vehicle mass and $I_z$ is the yaw inertia.
The contribution of the control input $\mathbf{u}$ (i.e., the control effectiveness matrix) is expressed as
\begin{equation}
  \label{eqa:g}
g(\mathbf{x},{r}) =
\begin{bmatrix}
0 & \tfrac{\cos\delta}{m} \\
0 & \tfrac{\sin\delta}{m} \\
0 & \tfrac{\ell_f \sin\delta}{I_z} \\
1 & 0
\end{bmatrix}.
\end{equation}
This formulation ensures that the steering rate $\dot{\delta}$ is directly controlled, while the longitudinal force $F_x$ is projected through the current steering angle $\delta$, influencing both lateral motion and yaw dynamics, assuming $F_x$ acts at the front axle along the wheel heading.

\subsubsection{\glslong{piml}}

Beyond purely analytical dynamics, a growing body of work explores \gls{piml} as a means to reconcile first-principles modeling with data-driven learning. The core objective is to leverage known physical structure, such as governing equations and kinematic constraints, while utilizing machine learning to capture unmodeled or highly nonlinear phenomena that analytical models often fail to represent.

Table~\ref{tab:physics_ml_comparison} provides a taxonomy of prominent \gls{piml} families. These range from \glspl{pinn}, which embed physical laws directly into the loss function as constraints for a data-driven \gls{pde}/\gls{ode} solver, to surrogate models that approximate expensive simulators, and hybrid residual approaches that learn corrections to a nominal physics model. Other categories include physics-regularized networks, Neural \glspl{ode} that integrate neural networks within continuous-time solvers, differentiable simulators for parameter identification, and operator-learning frameworks that approximate solution mappings.

These approaches differ fundamentally in their reliance on domain knowledge, their generalization capabilities, and the extent to which they preserve the underlying system structure. For instance, purely data-driven surrogate models maximize computational efficiency but often fail to generalize outside their training distribution. Conversely, operator learning can generalize across broad families of \glspl{pde} but typically demands large, diverse datasets.

In this work, we adopt a \emph{hybrid residual} strategy rooted in the dynamic bicycle model (Sec.~\ref{sec:bicycle}). Specifically, we design a structure-preserving architecture that retains the control-affine form of the dynamics while learning residual corrections from data. This approach balances physical interpretability with data-driven fidelity, ensuring the resulting model remains compatible with the safety-critical constraints required for geofence enforcement.

\subsection{Control models for geofence enforcement}

Given the vehicle dynamics defined above, we now describe two control models for geofencing intervention: a reactive braking-only model based on \gls{ttc}, and an active approach that exploits both braking and steering. 

\subsubsection{\glslong{ttc}}

The \gls{ttc} method represents a conservative, braking-only baseline. We reference the standard notion of time-to-collision from automotive practice (e.g. ISO~15623) and its origins and engineering use in collision-avoidance systems~\cite{iso15623,lee1976theory,van1994time}, while here evaluating safety via a max-braking rollout rather than a constant-relative-speed \gls{ttc} formula. This simulation of a fail-safe maneuver is conceptually related to formal verification methods that use reachability analysis to compute forward-reachable sets (often including designated fail-safe trajectories to a standstill) and verify whether a safe state can be maintained~\cite{althoff2014, pek2021failsafe}.

It determines whether a safe stop can be achieved from the current state by applying a fixed \gls{eb} control policy held constant during the rollout (zero steering rate, maximum-magnitude braking)
\(\mathbf{u}_{\text{EB}} = \begin{bmatrix} 0 \\ F_{x,\min} \end{bmatrix},\)
where $F_{x,\min}$ is the most negative admissible longitudinal force (maximum-magnitude braking).
A forward rollout is computed by numerically integrating the dynamics on a fixed grid \(t=0,0.1,\ldots,5.0\) [s] under $\mathbf{u}_{\text{EB}}$, terminating at the first index \(k_{\text{stop}}\) where the body-frame forward velocity \(v_x(k)\le 0\).
If there exists any sampled step $k\le k_{\text{stop}}$ such that $\sdf(\mathbf{p}_k)<-\varepsilon$ with $\mathbf{p}_k=[p_{x,k},p_{y,k}]^\top$ and $\varepsilon=0.5\,\text{m}$ (boundary tolerance), the current state is deemed unsafe.
This binary assessment captures whether an emergency stop is sufficient to guarantee containment. This approach is deliberately conservative and may lead to unnecessary stops or false interventions, a known issue with braking-only and other conservative geofencing methods that can trigger needless intervening actions or excessively restrict the operational area~\cite{thoren2024model}.
It is important to note that this ``braking-only rollout'' is closely aligned with classical \gls{ttc}-based safety evaluation~\cite{iso15623,lee1976theory,van1994time}, even though we implement it via a forward max-braking rollout rather than a closed-form constant-relative-speed \gls{ttc} formula.
In parallel, a separate line of modern work investigates the direct use of time-based indicators such as \gls{ttc} (or its inverse) as the safety constraint $h(x)$ within a \gls{cbf}-based optimization framework for longitudinal collision avoidance and traffic control~\cite{ZHAO2023104230}. These time-based \glspl{cbf} are particularly powerful for car-following scenarios, whereas our controller adopts the classical rollout as a conservative baseline and focuses on enforcing safety via a geometric signed-distance barrier $h(x)=\sdf(\pos)$, which is more naturally suited to arbitrary polygonal keep-in constraints.

\subsubsection{\glslong{dcbf}}

Let $\sdf(\pos)$ be the geometric signed distance (positive inside), and define the barrier $h(x):=\sdf(\pos)$ where $\pos=[p_x,p_y]^\top$ is extracted from the world-frame state $x$.
For continuous-time control-affine dynamics, a standard (exponential) \gls{cbf} condition enforces
\[
\dot{h}(x) + \kappa\, h(x) \;\ge\; 0,
\]
where $\kappa>0$ sets the convergence rate to the safe set.

While this condition defines safety, it does not specify the control law. The seminal work of \cite{ames_control_2017} established the Control Barrier Function-based Quadratic Program (CBF-QP) as the de facto framework for operationalizing this guarantee. In this paradigm, the CBF condition is formulated as an affine inequality constraint on the control input $\mathbf{u}$ and embedded in a real-time QP that mediates safety with a performance objective (e.g. a Control Lyapunov Function or a min-deviation term) \cite{ames_control_2017}.
Following the CBF–QP safety-filter template, and \emph{departing from the usual CLF-soft/CBF-hard convention in} \cite{ames_control_2017}, we adopt a \emph{heavily penalized CBF slack} in discrete time to preserve feasibility under preview linearization and model mismatch; when $\boldsymbol{\xi}=0$, we recover strict CBF enforcement.

While classical \glspl{cbf} are posed in continuous time, real implementations sample the dynamics at fixed intervals. We therefore adopt a discrete \gls{cbf} formulation that acts on the numerically integrated next-state map $x_{k+1}=\Phi(x_k,\mathbf{u}_k)$. We use the widely adopted discrete exponential surrogate:
\begin{equation}
h(x_{k+1}) \;\ge\; e^{-\kappa\Delta t}\, h(x_k) \;=: \beta_k,
\label{eq:disc-cbf}
\end{equation}
with $1-\gamma=e^{-\kappa\Delta t}$ (cf. the discrete exponential surrogate $\Delta h \ge -\gamma h$ in \cite{agrawal_discrete_2017}). With a fixed preview horizon $t_h$, we parameterize $\kappa=-\ln(1-\gamma)/t_h$ so the terminal target $\beta(t_h)$ used in Sec.~\ref{subsec:method-controller} is consistent with \eqref{eq:disc-cbf}.

\textit{Notation.} We use $h(x)=\sdf(\mathbf p)$ with safe set $\{x:h(x)\ge 0\}$; the discrete contraction factor is $\gamma\in[0,1)$ and the preview horizon is $t_h$; the terminal target is $\beta(t_h)=\max\{h_{\text{target}},\,h(x_0)e^{-\kappa t_h}\}$ with $\kappa=-\ln(1-\gamma)/t_h$.

\subsection{Numerical integration}
\label{sec:integration}
The vehicle dynamics are specified in continuous time as an ODE,
but controllers and safety checks operate in discrete time.
Numerical integration bridges this gap by approximating the evolution
$\dot{\mathbf{x}}=f(\mathbf{x},{r})+g(\mathbf{x},{r})\mathbf{u}$
with a discrete map $\mathbf{x}_{k+1}=\Phi(\mathbf{x}_k,\mathbf{u}_k)$ under \gls{zoh} inputs.
A \emph{rollout} is obtained by repeatedly applying this update, starting from an initial state, to generate a trajectory $\{x_0,x_1,\ldots,x_K\}$ over a horizon.

Depending on the task, integration must balance \emph{accuracy} (e.g. for training and validation)
and \emph{speed/stability} (e.g. for real-time safety checks).
To accommodate both needs we employ two complementary one-step schemes:

\paragraph{Classical \gls{rk4}.}
A fourth-order method that achieves high accuracy by averaging four slope evaluations per step.
It is used when fidelity and determinism are paramount, such as in training rollouts or stopping-distance safety checks.

\paragraph{Semi-implicit (symplectic) Euler.}
A first-order scheme, but qualitatively more stable than Forward Euler for mechanical states partitioned into positions and velocities.
Its low cost makes it well suited to the dense, short-horizon rollouts needed for predictive control barrier functions.
This trade-off mirrors practice in automotive simulation, where specially adapted semi-implicit Euler schemes are used to achieve stable, real-time-capable vehicle dynamics in \gls{hil} test stands~\cite{voegel1998realtime}.
Further background on the accuracy, stability, and error properties of these one-step schemes can be found in standard numerical analysis texts~\cite{hairer1993solving,butcher2016numerical,hairer2006geometric}.

\section{Related Work}
\label{sec:related}

Current geofencing strategies generally fall into geometric, reachability-based, or control-theoretic categories. Geometric methods, such as the Anticipatory Range Control (ARC) approach in \cite{thomas_geofencing_2024}, construct turning-circle envelopes to handle acute polygon corners, but, although originally developed for UAVs, typically assume constant turning radii that neglect vehicle-dynamics effects such as sideslip or tire slip when applied to ground vehicles. To guarantee containment under dynamic constraints, a Model Predictive Geofencing scheme based on \glspl{brt} is proposed in \cite{thoren2024model}, where the \gls{brt} for a given target set and dynamics is computed once offline and stored as a discretized tensor, yielding $\mathcal{O}(1)$ online execution but tying the safety guarantees to fixed model parameters (e.g., mass, surface friction, and braking limits). Closure-rate constraints for geofence violation prevention are derived in \cite{balachandran2018geofence} to ensure stopping distances are respected near a geofence, while an Explicit Reference Governor (ERG) is employed in \cite{hermand_constrained_2018} to adjust trajectory references upstream of the controller so that position and velocity constraints are never violated; however, such formally verified or ERG-based designs often require conservative braking profiles and linearized dynamics, which can limit operational efficiency or degrade fidelity at higher speeds.

On the other hand, real-time safety filters based on \glspl{cbf} and reachability have been developed for high-performance aerial vehicles. Implicit backup sets are used in conjunction with \gls{cbf}-based filtering in \cite{singletary_onboard_2022} to provide an onboard safety monitor for high-speed racing drones, and backstepping-based and high-order \glspl{cbf} for cascaded fixed-wing dynamics enabling simultaneous collision avoidance and geofencing for a Dubins-type aircraft model are designed in \cite{molnar_collision_2025}. Probabilistic extensions also exist; for example, \cite{bansal_hamilton-jacobi_2020} casts human motion prediction as a Hamilton--Jacobi reachability problem in a joint human--belief state space and computes belief-augmented forward reachable sets using grid-based HJ PDE solvers, although the focus there is on human motion rather than vehicle dynamics. Taken together, these approaches span geometric heuristics, precomputed reachable tubes and sets, and analytically derived barrier or governor conditions tailored to simplified models, highlighting a trade-off between computational tractability, model fidelity, and the ease with which safety constraints can be enforced in real time.

\paragraph{CBF controller design}
Digital controllers face the “inter-sample gap”: a system verified to be safe at discrete time steps may still violate constraints between samples as the physical system evolves continuously. Under \gls{zoh} implementation over a finite sampling interval, naively applying continuous-time \gls{cbf} designs can therefore create both inter-sample safety gaps and modeling gaps between exact and numerically integrated discrete-time maps~\cite{taylor_safety_2022,breeden_control_2022}. Extending \glspl{cbf} to discrete time on a general nonlinear next-state map $\Phi$ can also yield nonconvex synthesis problems; under exponential DT-CBF conditions, the problem becomes a QCQP~\cite{agrawal_discrete_2017}. For control-affine systems with mild block structure, DT-CBF invariance conditions that are affine in the input restore tractability and admit Boolean/piecewise-safe-set compositions via mixed-integer encodings~\cite{khajenejad_tractable_2021}. Sampled-data \gls{cbf} (SD-CBF) formulations explicitly address \gls{zoh} implementations:~\cite{taylor_safety_2022} formalize practical safety and show how SD-CBFs can be synthesized on approximate Runge--Kutta maps with consistency guarantees, while~\cite{breeden_control_2022} derive ZOH-aware, control-affine conditions that reduce conservativeness and embed cleanly in \glspl{qp}. Complementary zero-order formulations enforce safety directly on the numerically integrated next-sample map under \gls{zoh}, sidestepping Lie-derivative bookkeeping and accommodating high relative degree and state–input-dependent constraints~\cite{tan_zero-order_2025}, and high-order/adaptive DT-CBFs provide additional tools for feasibility under tightening bounds~\cite{xiong_discrete-time_2023}. For geofence enforcement under digital control, these discrete and sampled-data formulations identify how safety constraints should be posed on the numerically integrated map so that constraint sets remain invariant despite sampling and modeling errors.

Position-based safety constraints (such as keeping a vehicle within a geofence) typically have high relative degree: actuation affects acceleration or curvature, and position only changes after integrating velocity and acceleration. For such high-relative-degree constraints, continuous-time \gls{cbf} designs introduce additional structure so that the input appears in the barrier condition.
Early work used \emph{backstepping/dynamic extension} to handle relative degree $>1$ within a CBF–CLF–QP framework \cite{hsu_control_2015}. \emph{Exponential \glspl{cbf} (ECBFs)} provide a \emph{systematic, pole-placement} recipe to handle \emph{arbitrary} relative-degree constraints and embed them in \glspl{qp} \cite{nguyen_exponential_2016}. 
Alternatively, sampled-data approaches enforce barrier conditions on a previewed terminal state obtained by numerically integrating the dynamics over one sampling interval. By constraining the discrete barrier difference at that terminal state, these methods implicitly resolve high-relative-degree effects and can accommodate state–input-dependent constraints without recursive Lie-derivative constructions~\cite{tan_zero-order_2025}.

\paragraph{Geofence barrier parameterization and keep-in enforcement}
A fundamental distinction in safety filtering is the definition of the hazard. Two commonly adopted parameterizations are geometric-based and time-based. An extensive body of work uses time-based quantities such as time headway or time-to-collision as the core safety constraint $h(x)$ within \gls{cbf}-based longitudinal controllers for connected and automated vehicles~\cite{ZHAO2023104230}. However, time-based metrics are ill-posed for general 2D confinement, as the ``time to collision'' varies discontinuously with steering. Thus, spatial enforcement typically relies on geometric signed-distance barriers $h(x)=\sdf(\pos)$. Such time-based \glspl{cbf} are highly effective for one-dimensional car-following and traffic scenarios, but are less directly applicable to maintaining invariance within a general two-dimensional polygonal region. For geofencing and related 2D confinement problems, this motivates the use of signed-distance-based barriers that directly encode the margin to polygonal boundaries.

Greedy, single-step safety filters suffer from myopia, ''potentially steering the system into a dead end'' (inevitable collision) that is not immediately visible without a prediction horizon.
\emph{Predictive safety filters} use short-horizon optimization with a \emph{terminal safe set}. \emph{MPC–CBF} formulations enforce \emph{DT-CBF} constraints \emph{across the horizon}, powerful but requiring a \emph{finite-horizon} (generally nonlinear) program at each cycle \cite{zeng_safety-critical_2021}. \emph{Predictive Barrier Functions} introduce an \emph{always-feasible soft-constrained} auxiliary problem with a \emph{terminal CBF} and show its value function is itself a \emph{predictive CBF}, yielding a \emph{recovery mechanism} \cite{wabersich_predictive_2022}.

\paragraph{\gls{piml} models for vehicle dynamics and control}
Neural network vehicle models have been embedded in feedback control for high-performance driving. \cite{spielberg_neural_2019} replaces a simple physics-based model in a feedforward--feedback architecture with a neural model that takes a short history of past states and inputs as input, trained on real driving data, achieving human-competitive path tracking at the friction limit and implicitly adapting to changing road surfaces without explicit friction estimation. 
\cite{chrosniak2024} estimates tire, drivetrain, and inertia parameters of a single-track racecar model using a physics-constrained network with a guard layer that restricts coefficients to physically meaningful ranges, while~\cite{kabzan_learning-based_2019} use a sparse Gaussian-process residual on top of a nominal bicycle model inside an MPC controller, improving lap times and accounting for model uncertainty. \cite{jiang_high-accuracy_2021} and~\cite{miao2025residual} further develop residual-correction frameworks in which a deep encoder with an SVGP (DRF) or a Transformer (DyTR) learns trajectory- or state-prediction residuals on top of baseline vehicle models, substantially reducing prediction error, and~\cite{chen_lateral_2024} combine a mechanism-based estimator with an LSTM for lateral-velocity estimation that transfers across vehicles. \cite{kim_physics_2022} embed a Pacejka tire model and a dynamic bicycle model as differentiable layers inside a neural network, using latent tire-force features from the physics-embedded model to drive a risk-aware MPC that adapts to unknown friction. Overall, these works target high-fidelity prediction or racing performance, but their learned components typically act as generic residuals or parameter correctors and do not enforce an explicit control architecture, such as control-affine structures. This limitation hinders the design of verifiable safety-critical controllers, as the highly nonlinear dependency on the control input complicates the derivation of formal stability or safety guarantees.

\gls{piml} have also been proposed to explicitly adapt to control architectures. \cite{li_physics-informed_2024} replace an AGV kinematic model inside NMPC with a \gls{pinn} that enforces \gls{ode} residuals in the loss, and~\cite{antonelo_physics-informed_2024} introduce PINC, a \gls{pinn}-style architecture that supports control inputs and long-horizon simulation for nonlinear benchmark systems. In parallel, control-structure-aware neural identifiers explicitly align network structure with controller needs: CA-NNARX models are surrogate neural networks parameterized to be affine in the control input and come with $\delta$ISS conditions enabling efficient IMC design~\cite{xie2024}, and stable GRU models have been trained for system identification and embedded in nonlinear MPC with offset-free tracking guarantees~\cite{bonassi_nonlinear_2021}.

\section{Method}
\label{sec:method}
This section presents our geofencing pipeline. The design follows two principles: (i) a hybrid residual dynamics model that preserves a \emph{control-affine} structure to enable CBF-QP–style safety filtering, and (ii) a sampled-data, \emph{preview-discrete} \glslong{dcbf} controller that delivers minimal, stability-aware interventions. We first relate the background to the concrete geofencing task, then detail the model, the data-driven parameter refinement, and the controller, and finally summarize the full pipeline in algorithmic form.

\subsection{Overview}
\label{subsec:method-bridge}

The background (vehicle dynamics, barrier functions, and numerical integration) is instantiated for geofence enforcement via a two-stage architecture:

\noindent\textbf{Predictive layer (model-based preview).}
From the current world-frame state
\(
x=[p_x,p_y,\psi, v_{x,w}, v_{y,w}, \omega, \delta]^\top,
\)
we generate short-horizon rollouts under candidate constant inputs
\(
\mathbf{u}=[\dot\delta, F_x]^\top
\)
using the body-frame bicycle dynamics \( \dot{\mathbf{x}} = f(\mathbf{x},r) + g(\mathbf{x},r)\,\mathbf{u} \) with \glslong{zoh}. We integrate with a semi-implicit (symplectic) Euler preview scheme with a few substeps (for stability and speed); high-fidelity rollouts (e.g. for validation) use RK4 (cf. Sec.~\ref{sec:integration}). Body-frame accelerations are mapped to the world frame to update the state, and barriers are evaluated on the world position \(\pos=[p_x,p_y]^\top\).

The preview evaluates terminal barrier at the end of the horizon via the signed-distance barrier \(h(x)=\sdf(\pos)\)
(see Sec.~\ref{subsec:method-controller}), using the target \(\beta(t_h)\) from Eq.~\eqref{eq:beta-schedule}, consistent with Eq.~\eqref{eq:disc-cbf}.
To make the sensitivity well-posed near actuator limits, steering-rate sensitivities are computed by finite differences with \(\dot\delta\) steps \emph{clamped} to bounds, and the longitudinal-force column uses a bound–secant toward \emph{full braking} \(F_x=u_{\min}\).

\noindent\textbf{Control layer (minimal intervention).}
Given the preview, we enforce the terminal constraint $h(\Phi_{t_h}(x_k,\mathbf{u}_k)) \ge \beta(t_h)$ by solving a small 2-variable QP in \([\dot\delta, F_x]\) that minimally perturbs a nominal input \(\mathbf{u}^{\mathrm{nom}}\).
We linearize the terminal constraint about \(\mathbf u_k^{\mathrm{nom}}\) using the sensitivity matrix \(\mathbf J_k \approx \partial (h\!\circ\!\Phi_{t_h})/\partial \mathbf u\):
\[
h(\Phi_{t_h}(x_k,\mathbf u_k)) \;\approx\; h_{\text{nom}} + \mathbf J_k\,(\mathbf u_k-\mathbf u_k^{\mathrm{nom}}),
\]
where $h_{\text{nom}}:=h\!\big(\Phi_{t_h}(x_k,\mathbf u_k^{\mathrm{nom}})\big)$.
Including a slack variable $\boldsymbol{\xi}\!\ge\!0$ for soft-constraint feasibility, the constraint becomes:
\[
h_{\text{nom}} + \mathbf J_k\,(\mathbf u_k-\mathbf u_k^{\mathrm{nom}}) + \boldsymbol{\xi} \;\ge\; \beta(t_h).
\]
We write this in the standard affine form $\mathbf{A}\mathbf{u}_k + \boldsymbol{\xi} \;\ge\; \mathbf{b}$, with:
\[
\mathbf{A}=\mathbf J_k, \qquad
\mathbf{b}=\beta(t_h)-h_{\text{nom}}+\mathbf J_k\,\mathbf u_k^{\mathrm{nom}}.
\]
Minimal intervention is enforced by (i) an early exit when the terminal target is already satisfied (Algorithm~\ref{alg:runtime}), and (ii) an $\ell_2$-closest control to $\mathbf{u}^{\mathrm{nom}}$ subject to linearized safety and actuator box constraints (Eq.~\eqref{eqa:qp}). This QP instantiates the CBF-QP framework \cite{ames_control_2017} in a discrete-time, preview-terminal setting: unlike the common continuous-time setup (hard CBF, soft CLF), we use a proximal objective around $\mathbf{u}^{\mathrm{nom}}$ and a \emph{heavily penalized slack} on the CBF rows to preserve feasibility, together with actuator box constraints.

Only the control layer has actuator authority; the predictive layer proposes but does not actuate.

Compared with (i) TTC/\gls{eb}-only baselines that brake even when a gentle steer suffices, and (ii) continuous-time CBFs that require higher-order Lie derivatives and are brittle under sampling and saturation, our \emph{preview-discrete} \gls{dcbf} operates on the numerically integrated map, uses the target \(\beta(t_h)\), and builds saturation-aware Jacobians—improving feasibility and reducing unnecessary stops.

\subsection{Predictive layer: \gls{pcarnn} for modeling vehicle dynamics}
\label{subsec:method-model}

We retain a control–affine structure in the body frame (cf.~\ref{eqa:control-affine}), but $f$ and $g$ are represented by a combination of physics and neural networks.
More specifically, to preserve the control-affine form, \gls{pcarnn} separates learned corrections into drift and gain terms:
\begin{eqnarray}
f(\mathbf{x})&=&f_{\text{phys}}(\mathbf{x})+\Delta f_{\text{NN}}(\mathbf{x}),\\
g(\mathbf{x})&=&g_{\text{phys}}(\mathbf{x})+\Delta g_{\text{NN}}(\mathbf{x}),
\end{eqnarray}
where $f_{\text{phys}}, g_{\text{phys}}$ are the analytical bicycle terms, while the neural heads $\Delta f_{\text{NN}}$ and $\Delta g_{\text{NN}}$ learn structured residuals that capture tire nonlinearities, load transfer, and parametric drift.
We \emph{zero-initialize} $\Delta g_{NN}$ and bind the last row of $g_{\text{total}}$ to $[1~~0]$ so that $\dot{\delta}$ remains directly actuated and the model always stays control–affine in $(\dot{\delta},F_x)$.

\noindent\textbf{Novelty (dynamics).}
In the context of polygonal geofencing with sampled-data safety filters, we instantiate a physics-encoded, control–affine residual architecture with three key properties:
(i) it is \emph{hybrid residual}—data corrects an analytic dynamic-bicycle model instead of replacing it;
(ii) it is \emph{explicitly control–affine in} $\mathbf{u}=[\dot{\delta},F_x]^\top$, so that the safety constraint’s sensitivity $\mathbf{J}_k$ remains numerically stable and the resulting QP-based controller (Eq.~\eqref{eqa:qp}) stays as a small, two-variable problem that behaves reliably in the sampled-data \gls{cbf} setting~\cite{agrawal_discrete_2017,khajenejad_tractable_2021,tan_zero-order_2025};
and (iii) it is \emph{gracefully degradable}: setting $\Delta f_{\text{NN}}=\mathbf{0},\Delta g_{\text{NN}}=\mathbf{0}$ reverts to the physics baseline.
Unlike generic PIML vehicle models and black-box control-affine identifiers used in model-based control, this \gls{pcarnn} keeps an analytic dynamic-bicycle backbone with interpretable parameters and explicit steering-rate and braking channels, tailored to the geofence safety filter.\footnote{See Sec.~\ref{sec:related} for a detailed comparison with data-driven vehicle models and control-affine neural identifiers.}

While generic learned models (e.g. Neural \gls{ode}s) yield a nonlinear next-state map $\Phi(x,\mathbf{u})$ that can make safety filtering ill-conditioned or intractable, the proposed \gls{pcarnn} preserves separability of $f$ and $g$ (Eqs.~(6)--(7)), ensuring the terminal-barrier linearization (Eq.~\eqref{eqa:qp}) defines a small, well-posed \gls{qp} suitable for fast, tractable enforcement.

For integration, we use \gls{rk4} for high-fidelity training rollouts and semi-implicit Euler for fast preview inside the controller; positions and velocities are consistently mapped between body and world frames using $R(\psi)$.

Accurate physics terms depend on parameters $p$ (mass $m$, inertia $I_z$, axle distances $\ell_f,\ell_r$, tire stiffnesses $C_f,C_r$, friction $\mu$, etc.). We combine measurement, formula-based derivation, literature priors, and a targeted data-driven refinement loop that optimizes primarily the tire parameters while keeping the structure in Eq.~\eqref{eqa:control-affine} intact. This preserves control–affinity and improves preview accuracy near handling limits.

\begin{algorithm}[t]
\caption{Dynamic model calibration with data-driven parameter refinement}
\label{alg:calibration}
\small
\begin{algorithmic}[1]
\State \textbf{Inputs:} trajectory dataset $\mathcal{D}=\{(\mathbf{x}_t,\mathbf{u}_t,\dot{\mathbf{x}}_t)\}_{t=1}^N$; measurements (mass, geometry); tire priors $(C_f,C_r,C,E)$; training horizon $T$
\State \textbf{Outputs:} calibrated parameters $p^\star=\{C_f^\star,C_r^\star,C^\star,E^\star\}$
\vspace{0.2em}
\State \textbf{Initialize:} set nominal $(C_f,C_r,C,E)$ and measured geometry/mass to form $p$
\For{\textbf{epoch} $=1,2,\ldots$}
  \For{\textbf{mini-batch} $(\mathbf{x},\mathbf{u},\dot{\mathbf{x}})\subset\mathcal{D}$}
    \State \textit{Predict} $\;\dot{\mathbf{x}}_{\mathrm{pred}} = f_{\mathrm{phys}}(\mathbf{x};p) + g_{\mathrm{phys}}(\mathbf{x};p)\,\mathbf{u}$
    \State \textit{Loss (L2)} $\;\;\mathcal{L}=\big\|\dot{\mathbf{x}}_{\mathrm{pred}}-\dot{\mathbf{x}}\big\|_2^2$
    \State \textit{Update} apply Adam~\cite{kingma2015adam} to $\log C_f,\log C_r,\log C$ and to $E$
  \EndFor
  \State \textit{Validate} held-out rollouts of length $T$ (RK4); early-stop on forecast and barrier-prediction error
\EndFor
\State \textbf{return} $p^\star$
\end{algorithmic}
\end{algorithm}

\noindent\textbf{Novelty (calibration).}
The physics parameters and residual heads are \emph{co-trained}, but under architectural constraints that keep the dynamics in the control–affine form required by the safety filter.
This geofencing-oriented co-calibration improves preview accuracy (especially in tire-limited regimes) while preserving the structure needed for tractable, safety-filter–compatible control.

The neural network residual heads, $\Delta f_{\text{NN}}$ and $\Delta g_{\text{NN}}$, are implemented as feedforward MLPs that take the current body-frame dynamic state $\mathbf{x}$ as input. Following modern deep learning practice, we use the SiLU (Sigmoid-weighted Linear Unit) activation in the hidden layers~\cite{hendrycks2016gaussian,ramachandran_searching_2017}. The exact network architectures (depth, width) and whether parameters for $f$ and $g$ are shared or split are treated as hyperparameters; the specific configurations used in our experiments are reported in Sec.~\ref{sec:experiments}.

\subsection{Control layer: \glslong{dcbf} for geofence enforcement}
\label{subsec:method-controller}

\noindent\textit{(i) Terminal preview and schedule.}
We evaluate the barrier at a fixed preview time $\tau=t_h$ by integrating the bicycle model with semi-implicit Euler (constant $\mathbf{u}$) and $n_{\text{sub}}$ substeps:
$x(\tau)=\Phi_{\tau}(x_0,\mathbf{u})$.
We enforce a target schedule
\begin{equation}
\beta(\tau) \;=\; \max\!\big\{\,h_{\text{target}},\; h(x_0)\,e^{-\kappa\tau}\,\big\},
\kappa \;=\; \frac{-\ln(1-\gamma)}{t_h},
\label{eq:beta-schedule}
\end{equation}
so that $h(x(\tau))\ge \beta(\tau)$ guarantees a monotonic approach toward a positive margin $h_{\text{target}}$ (cf.\ the discrete exponential DCLF–DCBF surrogate $h_{k+1}\!\ge\!(1-\gamma)h_k$ in \cite{agrawal_discrete_2017}).
In all experiments, unless noted otherwise, we use \(t_h=0.30\) s, \(n_{\text{sub}}=3\), \(\varepsilon_{\dot\delta}=0.25~\mathrm{rad/s}\), a longitudinal bound–secant toward full braking, \(\gamma=0.4\), and clip sensitivity magnitudes at \(\Gamma_{\text{clip}}=10^6\).

\paragraph{On high relative degree.}
By enforcing the barrier at a finite preview time $\tau$, the current input $\mathbf{u}$ directly influences the terminal state $x(\tau)$ under zero-order hold, and thus $h\big(x(\tau)\big)$.
Classical continuous-time \gls{cbf} constructions handle high-relative-degree position constraints by introducing chains of Lie derivatives via backstepping/dynamic extension~\cite{hsu_control_2015} or pole-placement formulations such as Exponential \glspl{cbf}~\cite{nguyen_exponential_2016}.
In contrast, we follow the sampled-data / zero-order \gls{cbf} viewpoint~\cite{agrawal_discrete_2017,taylor_safety_2022,breeden_control_2022,tan_zero-order_2025}: rather than enforcing conditions on continuous-time derivatives, we act directly on the numerically integrated preview map $h \circ \Phi_{\tau}$ and impose a discrete-time inequality on this map.
This avoids recursive high-order \gls{cbf} constructions and yields a terminal barrier condition that is \emph{locally} well-approximated as affine in $\mathbf{u}$ when combined with our control-affine \gls{pcarnn} dynamics.
The resulting constraint enters the \gls{qp} as a single linear row (Eq.~\eqref{eqa:qp}), with the polygonal signed-distance barrier, its speed-aware extension, and the saturation-aware sensitivities from the preview-rollout linearization tailoring the sampled-data condition to the polygonal geofencing and actuator-limited setting.

\noindent\textit{(ii) Polygonal SDF with inward normal.}
Let $\sdf(\pos)$ be the signed distance to a simple polygon \(\Omega\subset\mathbb{R}^2\), positive inside.
The inward unit normal is \(\mathbf{n}_{\text{in}}(\pos)=\nabla_{\pos}\sdf(\pos)\).
Since the state barrier is \(h(x)=\sdf(\pos)\) with \(\pos=[p_x,p_y]^\top\) and \(x=[p_x,p_y,\psi,v_{x,w},v_{y,w},\omega,\delta]^\top\),
the state-gradient is
\[
  \nabla_x h(x)
  =\big[\;\nabla_{\pos}\sdf(\pos)^\top\;\;\;\mathbf{0}_{1\times 5}\;\big].
\]

\noindent\textit{(iii) Input-space linearization by preview rollouts.}
We build the \emph{barrier sensitivity matrix} $\mathbf{J}=[\,J_{h,\dot\delta}\;\;J_{h,F_x}\,]$ at the preview time $\tau=t_h$ using short, zero-order-hold rollouts of the numerically integrated map $\Phi_{\tau}$. Let $\mathbf{u}_{\text{nom}}=[\dot\delta_{\text{nom}},\,F_{x,\text{nom}}]^\top$. Define the elementwise clipping operator $\mathrm{clip}(a; a_{\min}, a_{\max})=\min\{\max\{a,a_{\min}\},a_{\max}\}$ and the steering perturbations

\[
\dot\delta_{\pm}
\;=\;
\mathrm{clip}\!\big(\dot\delta_{\text{nom}}\pm \varepsilon_{\dot\delta};\;\dot\delta_{\min},\dot\delta_{\max}\big),
\qquad
\mathbf{u}_{\pm}=\big[\dot\delta_{\pm},\,F_{x,\text{nom}}\big]^\top,
\]
together with a full-braking input $\mathbf{u}_{\text{brk}}=[\dot\delta_{\text{nom}},\,F_{x,\min}]^\top$.

To capture the nonlocal safety effect of decisive braking, we use a bound--secant for the longitudinal entry:
\begin{equation}
\label{eq:J_Fx_secant_updated}
J_{h,F_x}
\;\approx\;
\frac{\,h\!\big(\Phi_{\tau}(x_0,\mathbf{u}_{\text{brk}})\big)
-
h\!\big(\Phi_{\tau}(x_0,\mathbf{u}_{\text{nom}})\big)\,}
{\,F_{x,\min}-F_{x,\text{nom}}\,}.
\end{equation}

This ``bound--secant'' approach (Eq.~\eqref{eq:J_Fx_secant_updated}) intentionally deviates from a standard local tangent approximation. It is a pragmatic heuristic designed to address the well-known challenges that actuator saturation poses for optimization-based controllers, which can lead to QP infeasibility or performance degradation when the linearized model's predictions diverge from the system's true capabilities~\cite{decastro20219}. Such saturation-induced infeasibilities are a known challenge in control-allocation formulations~\cite{JOHANSEN20131087}. By using a secant to the saturation limit $F_{x,\min}$, we provide the safety-filter QP with a more conservative, nonlocal approximation of the true control authority available, improving feasibility and ensuring that the ``true safety effect of hard braking'' is reflected in the optimization.

We first attempt a central difference using the clamped perturbations $\dot\delta_{\pm}$. If the central-difference interval collapses (i.e., $\dot\delta_{+}=\dot\delta_{-}$ because both perturbations clamp to the \emph{same} bound), we revert to a one-sided secant from the nominal input to that active bound (cases below):

\noindent{If $\dot\delta_{+}>\dot\delta_{-}$,}
\small{\begin{eqnarray}
J_{h,\dot\delta}
&\approx&
\frac{\,h\!\big(\Phi_{\tau}(x_0,\mathbf{u}_{+})\big)-h\!\big(\Phi_{\tau}(x_0,\mathbf{u}_{-})\big)\,}
{\,\dot\delta_{+}-\dot\delta_{-}\,}.
\label{eq:J_deldot_central_updated}
\end{eqnarray}}

\noindent{If $\dot\delta_{+}=\dot\delta_{-}=\dot\delta_{\min}$,}
\small{\begin{eqnarray}
J_{h,\dot\delta}
\approx
\frac{h\!\big(\Phi_{\tau}(x_0,[\dot\delta_{\text{nom}},\,F_{x,\text{nom}}]^\top)\big)
      -h\!\big(\Phi_{\tau}(x_0,[\dot\delta_{-},\,F_{x,\text{nom}}]^\top)\big)}
{\dot\delta_{\text{nom}}-\dot\delta_{-}},
\label{eq:J_deldot_one_sided_updated}
\end{eqnarray}}

\noindent{If $\dot\delta_{+}=\dot\delta_{-}=\dot\delta_{\max}$,}
\small{\begin{eqnarray}
J_{h,\dot\delta}
\approx
\frac{h\!\big(\Phi_{\tau}(x_0,[\dot\delta_{+},\,F_{x,\text{nom}}]^\top)\big)
      -h\!\big(\Phi_{\tau}(x_0,[\dot\delta_{\text{nom}},\,F_{x,\text{nom}}]^\top)\big)}
{\dot\delta_{+}-\dot\delta_{\text{nom}}}.
\label{eq:J_deldot_one_sided_max_updated}
\end{eqnarray}}

For conditioning, we clip derivative magnitudes:
\[
J_{h,\dot\delta}\leftarrow \mathrm{clip}(J_{h,\dot\delta};-\Gamma_{\text{clip}},\Gamma_{\text{clip}}),
\qquad
J_{h,F_x}\leftarrow \mathrm{clip}(J_{h,F_x};-\Gamma_{\text{clip}},\Gamma_{\text{clip}})
\]
with a fixed $\Gamma_{\text{clip}}>0$.
With the row for the terminal barrier and bounds stacked, we solve the minimal-deviation \gls{qp}:
\paragraph{QP instantiation.}
In our implementation we solve the softened CBF-QP in scaled solver variables $z=[\,v;\,\boldsymbol{\xi}\,]$, with $v=S\,\mathbf{u}$ and $S=\mathrm{diag}(1,10^{-3})$:
\[
\begin{aligned}
\min_{v,\,\boldsymbol{\xi}\ge 0}\quad
& \tfrac12\|v-v_{\mathrm{nom}}\|_{\Lambda}^2 \;+\; \tfrac12\,\rho_{\mathrm{slack}}\|\boldsymbol{\xi}\|_2^2,\\
\text{s.t.}\quad
& A\,S^{-1}v \;+\; \boldsymbol{\xi} \;\ge\; b, \qquad
 S\,\mathbf{u}_{\min} \;\le\; v \;\le\; S\,\mathbf{u}_{\max},
\end{aligned}
\]
where $v_{\mathrm{nom}}=S\,\mathbf{u}^{\mathrm{nom}}$, $\Lambda=\mathrm{diag}(\lambda_{\dot\delta},\lambda_{F_x})$, and $\rho_{\mathrm{slack}}\!\gg\!0$. We log $\|\boldsymbol{\xi}\|_{\infty}$ to monitor any residual violation; when $\boldsymbol{\xi}=0$ the constraints are met strictly.

\noindent\textit{Weight/scaling neutrality.} The relative influence of steering vs.\ braking in the proximal term is determined by the combination of variable scaling $S$ and solver weights $\Lambda$. Choosing $(S,\Lambda)$ is equivalent to choosing $W=S^{\top}\Lambda S$ in the unscaled form; i.e., any “preference” is a modeling choice rather than an inherent bias of the method. \emph{In the original (unscaled) variables we write:}
\begin{eqnarray}
  \label{eqa:qp}
\min_{\mathbf{u},\,\boldsymbol{\xi}\ge 0}
& & \tfrac12\|\mathbf{u}-\mathbf{u}^{\text{nom}}\|_W^2
     + \tfrac12\rho\|\boldsymbol{\xi}\|_2^2 \nonumber \\
\text{s.t.}
& & \mathbf{A}\mathbf{u} + \boldsymbol{\xi} \;\ge\; \mathbf{b}, \nonumber \\
& & \mathbf{u}_{\min} \;\le\; \mathbf{u} \;\le\; \mathbf{u}_{\max}\,,
\end{eqnarray}
where \(\mathbf{A}\) stacks the linearized barrier sensitivities (e.g. the row from \(h\) at \(\tau=t_h\)) and \(\mathbf{b}\) stacks the corresponding right-hand sides (e.g. \(\beta(t_h)-h_{\text{nom}}+\mathbf J\,\mathbf u^{\text{nom}}\)). Actuator bounds \(\mathbf{u}_{\min}\le\mathbf{u}\le\mathbf{u}_{\max}\) are retrieved online from the simulator, which encodes state-dependent steering-rate limits near mechanical stops and speed-dependent longitudinal force limits.

The above linearization acts on the \emph{previewed, sampled-data} terminal map $h\!\circ\!\Phi_{\tau}$ obtained by numerical integration under ZOH. This aligns with the sampled-data CBF viewpoint: it directly enforces a discrete barrier difference on the next-sample map, avoiding Lie-derivative bookkeeping (cf.\ the zero-order perspective \cite{tan_zero-order_2025}) and is consistent with SD-CBF formulations that justify designing safety filters on \emph{approximate} discrete-time models (e.g. Runge–Kutta) with practical-safety guarantees \cite{taylor_safety_2022}. In parallel, ZOH-aware CBF conditions for sampled-data systems provide control-affine, QP-enforceable constraints that reduce conservativeness relative to earlier emulation-style margins \cite{breeden_control_2022}. Our finite-horizon, terminal enforcement follows this sampled-data logic while adding saturation-aware sensitivities and a slack-penalized feasibility mechanism for robustness to model mismatch.

\noindent\textbf{Novelty (controller).}
Within the sampled-data / zero-order \gls{cbf} framework~\cite{agrawal_discrete_2017,taylor_safety_2022,breeden_control_2022,tan_zero-order_2025}, our controller makes three design contributions tailored to polygonal geofencing and actuator-limited ground vehicles:
(i) it instantiates a finite-horizon, preview-terminal barrier condition on the polygonal signed-distance map $h\circ\Phi_{\tau}$, using a geofence-specific schedule $\beta(\tau)$ and the learned \gls{pcarnn} dynamics so that the constraint enters the QP as a single linear row tailored to the keep-in constraint;
(ii) it \emph{introduces a simple ``bound--secant'' linearization heuristic} near the full-braking limit to improve \gls{qp} feasibility under actuator saturation, informed by control-allocation perspectives~\cite{JOHANSEN20131087,decastro20219};
and (iii) it \emph{solves a scaled minimal-deviation \gls{qp} with an early-exit condition} that avoids solving the \gls{qp} when the nominal control is already safe, thereby reducing both intervention and computational load.
Taken together, the preview-terminal enforcement on $h\circ\Phi_{\tau}$, the saturation-aware bound--secant sensitivities, and the scaled minimal-deviation \gls{qp} with early exit yield a single, tractable safety filter specialized to polygonal geofencing with actuator-limited, learned vehicle dynamics.

\subsection{End-to-end runtime pipeline}
\label{subsec:method-pipeline}

\begin{figure}
    \centering
    \resizebox{0.5\textwidth}{!}{\input{graph_system}}
    \caption{\gls{pcarnn-dcbf} geofencing architecture.}
    \label{fig:safety-qp}
\end{figure}
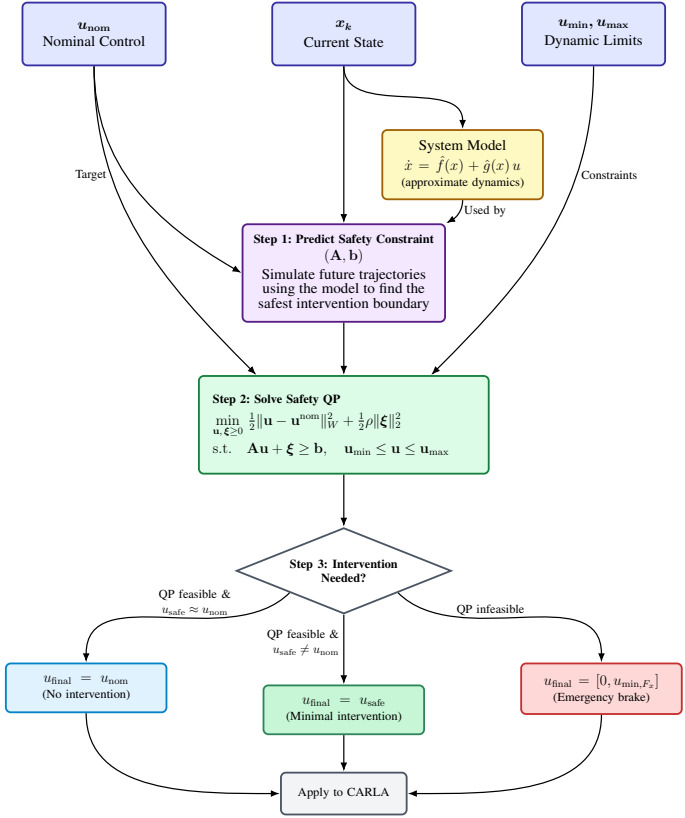

The end-to-end geofencing algorithm is presented in Algorithm~\ref{alg:runtime} and the overall architecture is shown in Figure~\ref{fig:safety-qp}.
If the \gls{qp} is infeasible (line 12), the controller is designed to revert to a predefined fallback policy, commanding a full emergency brake $\mathbf{u}_{\text{EB}}$. 
This provides a simple, conservative fail-safe: when the safety filter cannot find a minimally corrective control input that guarantees constraint satisfaction, the vehicle is brought to a stop inside the current geofence.
\begin{algorithm}[!t]
\caption{Runtime geofencing with \gls{pcarnn-dcbf}}
\label{alg:runtime}
\small
\begin{algorithmic}[1]
\State \textbf{Inputs:} current state $x_0$, polygonal geofence $\Omega$, nominal input $\mathbf{u}^{\text{nom}}$, model $(p^\star,\theta^\star)$, bounds $\mathbf{u}{\min},\mathbf{u}{\max}$, horizon $\tau$, substeps $n_{\text{sub}}$, targets $(h_{\text{target}},\kappa)$ [$\kappa$]
\State \textbf{Output:} commanded input $\mathbf{u}^\star$
\vspace{0.2em}
\State propagate preview under $\mathbf{u}^{\text{nom}}$: $x(\tau)\leftarrow\Phi_{\tau}(x_0,\mathbf{u}^{\text{nom}})$
\State evaluate terminal barrier $h$ and schedule $\beta(\tau)$
\If{$h(x(\tau))\ge\beta(\tau)$}
\State \textbf{return} $\mathbf{u}^{\text{nom}}$ \quad \textit{(early exit: no intervention)}
\EndIf
\State build Jacobian $\mathbf{J}$ by short rollouts:
\Statex \quad \textbf{(a)} central differences in $\dot{\delta}$ around $\mathbf{u}^{\text{nom}}$
\Statex \quad \textbf{(b)} bound–secant in $F_x$ between $\mathbf{u}^{\text{nom}}$ and $F_{x,\min}$
\State assemble linearized constraint $(\mathbf{A},\mathbf{b})$ for $h$
\State solve \gls{qp} Eq.~\eqref{eqa:qp} with bounds $\mathbf{u}{\min}\le\mathbf{u}\le\mathbf{u}{\max}$
\If{\gls{qp} feasible}
\State \textbf{return} $\mathbf{u}^\star$ \textit{(minimal correction; pass to on-board motion controller)}
\Else
\State \textbf{return} $\mathbf{u}{\text{EB}}=[0,,F{x,\min}]^\top$ \textit{(safety fallback)}
\EndIf
\end{algorithmic}
\end{algorithm}

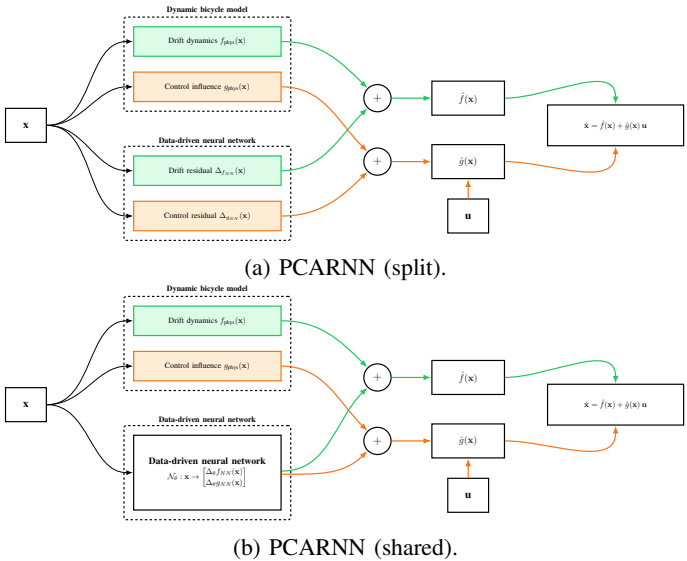
\begin{figure}[ht]
    \centering
    \begin{subfigure}{0.5\textwidth}
        \centering
        \resizebox{\textwidth}{!}{\input{graph_ca_model_split}}
        \caption{\gls{pcarnn} (split).}
        \label{fig:ca-model-split}
    \end{subfigure}
    \hfill
    \begin{subfigure}{0.5\textwidth}
        \centering
        \resizebox{\textwidth}{!}{\input{graph_ca_model_shared}}
        \caption{\gls{pcarnn} (shared).}
        \label{fig:ca-model-shared}
    \end{subfigure}
    \caption{System model: \gls{pcarnn} split model and \gls{pcarnn} shared model.}
    \label{fig:ca-model-comparison}
\end{figure}

\section{Experiments and Results}
\label{sec:experiments}

\subsection{Experimental Setup}
\subsubsection{Trajectory Creation in \gls{carla}}
All training and evaluation trajectories were generated using the \gls{carla} simulator (version 0.9.15) with a custom Python interface for high-fidelity control and state extraction~\cite{dosovitskiy2017carla}.
Each trajectory was constructed to ensure kinematic consistency between position, velocity, and acceleration measurements and to capture realistic dynamic behavior across different operating regimes.
Further details on the simulator configuration, dataset generation, and test scenario labeling are provided in Appendix~\ref{sec:appendix_carla}.

\subsubsection{Operating Regimes and Test Scenarios}
The distribution of safe and unsafe cases differs across both vehicles and operating regimes. A summary can be found in Table~\ref{tab:regime_summary}.
These trajectories were designed to evaluate how each model handles scenarios of varying difficulty, defined by combinations of vehicle speed and steering curvature.

In general, \textbf{low-speed-straight} regimes represent the easiest operating conditions.
Vehicle dynamics are slow, lateral drift is minimal, and small corrective inputs are sufficient to maintain containment within the geofence.
Similarly, \textbf{low-speed-sharp} trajectories, while involving higher steering inputs, remain relatively easy to control due to limited kinetic energy and lower momentum, which provide larger control margins and longer reaction times.

By contrast, \textbf{high-speed-straight} regimes are among the most challenging.
Even small deviations from the boundary can quickly accumulate due to limited lateral authority and high inertial forces, leaving little room for corrective steering or braking.
Finally, \textbf{high-speed-sharp} trajectories pose compounded difficulties: high curvature demands rapid and precise steering control, while increased velocity amplifies centrifugal forces and narrows the feasible control envelope.
These scenarios are thus critical for evaluating the controller’s responsiveness, stability, and ability to respect safety constraints under tight dynamic coupling between steering and braking.

Overall, this stratification of test scenarios enables a systematic assessment of how well each model generalizes across operating conditions that vary in control difficulty and physical constraints.
\begin{table}[!ht]
\centering
\caption{\label{tab:regime_summary}%
Counts of safe and unsafe scenarios by operating regime for each vehicle.
Regimes are defined using thresholds on vehicle speed ($v_x=8.0$\,m/s) and steering input ($\delta=0.35$\,rad).}
\resizebox{0.5\textwidth}{!}{
\begin{tabular}{l l r r r}
\toprule
\textbf{Vehicle} & \textbf{Operating Regime} & \textbf{Safe} & \textbf{Unsafe} & \textbf{Total} \\
\midrule
\multirow{4}{*}{Lincoln MKZ}
 & Low-Straight & 49 & 44 & 93 \\
 & Low-Sharp    & 65 & 28 & 93 \\
 & High-Straight & 17 & 18 & 35 \\
 & High-Sharp   & 29 &  8 & 37 \\
\cmidrule(lr){2-5}
 & \textbf{Total} & \textbf{160} & \textbf{98} & \textbf{258} \\
\midrule
\multirow{4}{*}{Audi E-tron}
 & Low-Straight & 63 & 58 & 121 \\
 & Low-Sharp    & 98 & 16 & 114 \\
 & High-Straight & 48 & 49 & 97 \\
 & High-Sharp   & 43 & 22 & 65 \\
\cmidrule(lr){2-5}
 & \textbf{Total} & \textbf{252} & \textbf{145} & \textbf{397} \\
\bottomrule
\end{tabular}
}
\end{table}

\subsubsection{Models for Benchmarking}

To benchmark different models and evaluate the effectiveness of the \gls{pcarnn-dcbf}, we implemented and tested various \{\texttt{model}\}-\gls{dcbf} architectures, where 
\begin{center}
\texttt{model}$\in$\{\texttt{Dynamic Bicycle Model}, \texttt{Neural \gls{ode}}, \texttt{Hybrid Residual Model}, \texttt{\gls{pcarnn} (shared-network)}, \texttt{\gls{pcarnn} (split-network)}\}.
\end{center}
\paragraph{Dynamic Bicycle Model}
The analytical baseline follows the kinematic bicycle formulation described in Section~\ref{sec:bicycle}.
For comparison, we also include the \emph{Bicycle Ackermann} variant (adopted in \cite{thoren2024model}). In this model, the front axle follows classical Ackermann steering: in a steady turn, the inner and outer front wheels are oriented so that their wheel axes intersect at a common centre of rotation, with the inner wheel commanded to a larger steer angle so that both wheels roll on circular paths without lateral slip in low-speed manoeuvres~\cite{corominas2014fourwheel,verma2019_ackermann}. An equivalent single steering angle~$\delta$ for the bicycle abstraction is then computed from the inner and outer wheel angles using this Ackermann relation, yielding a more geometrically accurate low-speed kinematic baseline.

\paragraph{Neural \gls{ode}}
A neural network $NN_\theta(\mathbf{x}, \mathbf{u})$ directly models the system dynamics by predicting the state derivative:
\[
\dot{\mathbf{x}} = NN_\theta(\mathbf{x}, \mathbf{u}).
\]
This formulation maximizes expressiveness but abandons the control-affine structure, providing no guarantee of separability between system dynamics and control input.
\begin{table*}[!ht]
\centering
\caption{Summary of best-performing variants (highest \gls{cf1}) from each model group (under braking + steering settings) for both vehicles. One small ($\sim$50K) and one large ($\sim$150K) configuration are selected per variant type. Lower is better for \gls{fpr} and \gls{mcd+}, higher is better for \gls{cf1}.}
\resizebox{1\textwidth}{!}{
\begin{tabular}{
  p{1.5cm}  
  p{2.5cm}  
  p{1.8cm}  
  p{4.5cm}  
  p{2.1cm}  
  p{2.1cm}  
  p{2.1cm}  
  p{2.1cm}  
}
\toprule
\textbf{Vehicle} & \textbf{Model Variant} & \textbf{Controller} & \textbf{Size ($f$, $g$)} & \textbf{Params} & \textbf{CF1}${\uparrow}$ & \textbf{FPR}${\downarrow}$ & \textbf{MCD}$^{+}{\downarrow}$ \\
\midrule

\multirow{5}{*}{Audi E-tron}
  & PCARNN (shared) & DCBF & $4\times128$ / $5\times192$ & 51.3K / 150K & 0.915 / {0.918} & 0.082 / 0.077 & \textbf{0.438} / 0.498 \\
  & PCARNN (split) & DCBF & $(3\times90, 4\times104)$ / $(5\times135, 5\times135)$ & 51K / 149K & \underline{0.928} / \textbf{0.935} & 0.087 / \underline{0.066} & 0.672 / \underline{0.470} \\
  & Residual & DCBF & $4\times128$ / $5\times192$ & 50.9K / 150K & 0.906 / 0.913 & 0.082 / \textbf{0.060} & 0.760 / 0.489 \\
  & Neural ODE & DCBF & $4\times128$ / $5\times192$ & 50.9K / 150K & 0.910 / 0.906 & 0.077 / {0.082} & {0.734} / {0.746} \\
  & Bicycle & DCBF & N/A & N/A & {0.850} & 0.077 & 0.617 \\
  & Bicycle Ackermann & DCBF & N/A & N/A & {0.896} & 0.120 & 1.080 \\
\midrule

\multirow{5}{*}{Lincoln MKZ}
  & PCARNN (shared) & DCBF & $4\times128$ / $5\times192$ & 51.3K / 150K & \textbf{0.922} / \underline{0.913} & \underline{0.050} / 0.062 & 0.356 / 0.569 \\
  & PCARNN (split) & DCBF & $(3\times90, 4\times104)$ / $(4\times128, 5\times156)$ & 51K / 150K & {0.908} / \underline{0.913} & 0.069 / {0.062} & 0.647 / {0.535} \\
  & Residual & DCBF & $4\times128$ / $5\times192$ & 50.9K / 150K & 0.737 / {0.748} & \textbf{0.044} / 0.062 & \textbf{0.293} / \underline{0.341} \\
  & Neural ODE & DCBF & $4\times128$ / $5\times192$ & 50.9K / 150K & 0.718 / {0.771} & 0.062 / 0.056 & 0.344 / 0.359 \\
  & Bicycle & DCBF & N/A & N/A & {0.832} & \textbf{0.044} & 0.457 \\
  & Bicycle Ackermann & DCBF & N/A & N/A & {0.905} & 0.088 & 0.999 \\

\bottomrule
\end{tabular}}
\end{table*}

\paragraph{Hybrid Residual Model}
The residual model augments the analytical dynamics with a learned correction term:
\[
\dot{\mathbf{x}} = f_{\text{phys}}(\mathbf{x}) + g_{\text{phys}}(\mathbf{x})\mathbf{u} + \Delta_\theta(\mathbf{x}, \mathbf{u}),
\]
where $\Delta_\theta$ is a neural residual. This approach anchors learning to known physics while improving expressiveness. However, the addition of $\Delta_\theta$ breaks the strictly control-affine form since it may depend nonlinearly on $\mathbf{u}$.

\paragraph{\gls{pcarnn} (shared-network)}
In this formulation, we learn residuals for both $f$ and $g$ using a shared neural network structure.
A single network outputs both the residual components for $f$ and $g$,
with the first part of the output $\Delta_{{\theta}}f(\mathbf{x})$ corresponding to the drift term and the second $\Delta_{{\theta}}g(\mathbf{x})$ to the control matrix.
The final dynamics remain control-affine:
\[
\dot{\mathbf{x}} = (f_{\text{phys}}(\mathbf{x}) + \Delta_{{\theta}}f(\mathbf{x})) + (g_{\text{phys}}(\mathbf{x}) + \Delta_{{\theta}}g(\mathbf{x}))\mathbf{u}.
\]
We evaluate both small ($\sim$51.3K parameters) and large ($\sim$150K parameters) configurations.

\paragraph{\gls{pcarnn} (split-network)}
This variant employs two independent neural networks for $\Delta f_{NN}$ and $\Delta g_{NN}$,
allowing greater specialization for the drift and control components, respectively.
The resulting model preserves the control-affine structure:
\[
\dot{\mathbf{x}} = (f_{\text{phys}}(\mathbf{x}) + \Delta f_{NN}(\mathbf{x})) + (g_{\text{phys}}(\mathbf{x}) + \Delta g_{NN}(\mathbf{x}))\mathbf{u}.
\]
As with the shared-network variant, we evaluate two model sizes, approximately 51K and 150K parameters, corresponding to compact and large-scale configurations.

\subsubsection{Network Architectures}
The $\Delta f_{\text{NN}}$ and $\Delta g_{\text{NN}}$ components were implemented as standard feedforward neural networks. We experimented with several architectures to analyze the impact of model capacity and parameter sharing, including:
\begin{itemize}
    \item \textbf{Shared-network:} A single MLP backbone that outputs corrections for both $f$ and $g$.
    \item \textbf{Split-network:} Two independent MLPs, one for $\Delta f_{\text{NN}}$ and one for $\Delta g_{\text{NN}}$.
\end{itemize}
These shared and split control-affine architectures are summarized schematically in Figure~\ref{fig:ca-model-comparison}.
For both variants, we tested two network sizes, referred to as ``small'' ($\sim$50K parameters) and ``large'' ($\sim$150K parameters). For example, a large configuration (e.g. \texttt{big\_5x192}) consists of 5 hidden layers with 192 nodes each, as shown in our ablation studies (Table~\ref{tab:hyperparameter}). Following modern practice, we use the SiLU activation in hidden layers~\cite{hendrycks2016gaussian,ramachandran_searching_2017}. All models were trained with the AdamW optimizer~\cite{loshchilov_decoupled_2019} and a cosine-annealing learning-rate schedule~\cite{loshchilov_sgdr_2017}.
 In addition, we apply spectral normalization to the linear layers of the residual networks to control their operator~$2$-norms and thereby bound the Lipschitz gains of the learned corrections~\cite{miyato2018spectral}. In the split \gls{pcarnn}, spectral normalization is applied only to the $\Delta g_{\text{NN}}$ network, while in the shared \gls{pcarnn} a single spectrally normalized backbone produces both $\Delta f_{\text{NN}}(\mathbf{x})$ and $\Delta g_{\text{NN}}(\mathbf{x})$.

\subsection{Evaluation metrics}
The effectiveness of geofence enforcement is evaluated based on three key criteria:
(1) it should successfully \emph{detect} potential breaches and \emph{prevent} the vehicle from breaching the geofence once an intervention is applied (high intervention success),
(2) the controller should \emph{accurately detect} only when an intervention is necessary (low false alarm),
and (3) the \emph{intervention should be minimal}, avoiding unnecessary or excessive control actions (minimal intervention).
To capture these aspects quantitatively, we employ three complementary metrics: \gls{fpr}, \gls{cf1}, and \gls{mcd+}.
At each control cycle we cast the decision as a binary classification.
The predicted label is \emph{positive} if the controller issues any intervention (brake and/or steer), otherwise \emph{negative}.
The ground-truth label is \emph{positive} if the nominal, no-intervention rollout would violate the geofence within the lookahead horizon, otherwise \emph{negative}.
Let $TP, FP, TN, FN$ denote the corresponding counts, and let $CF$ be the number of \emph{containment failures}, i.e. ground-truth positive cases where the controller \emph{did} intervene but the geofence was still breached downstream (``try to intervene but failed'').
\subsubsection{\textbf{\gls{cf1}}}
The Containment F1 score augments the standard F1 with a penalty for failed containments among required interventions.
Define the success ratio among true positives as
\[
s=\frac{TP-CF}{TP}, \qquad \text{(set } s=0 \text{ if } TP=0).
\]
Then
\[
\mathrm{\gls{cf1}} \;=\; F1 \cdot s \;=\; F1 \cdot \frac{TP-CF}{TP}.
\]
Higher values indicate better containment performance, rewarding both correct intervention timing and effective containment. 
This is the main evaluation metric.
\subsubsection{\textbf{\gls{fpr}}}
The false positive rate measures unnecessary interventions:
\[
\mathrm{\gls{fpr}} \;=\; \frac{FP}{FP+TN}.
\]
Lower values indicate fewer unnecessary interventions.

\subsubsection{\textbf{\gls{mcd+}}}
The median containment distance measures how close trajectories come to the geofence boundary under the closed-loop controller.
For each episode, we compute the minimum signed distance to the polygon over time,
\[
d_{\min} \;=\; \min_t \; \sdf\!\big(\pos(t)\big),
\]
where $\sdf(\cdot)>0$ when the vehicle is inside the fence and $\sdf(\cdot)<0$ when it is outside.
We then report the median of these minimum distances across all episodes:
\[
\mathrm{\gls{mcd}} \;=\; \mathrm{median}\big(\{d_{\min}^{(i)}\}_i\big)\;\;[\text{m}].
\]
A positive $\mathrm{\gls{mcd}}$ indicates that the vehicle remains within the geofence, where smaller positive values correspond to minimal but safe containment margins.
A negative $\mathrm{\gls{mcd}}$ indicates boundary violation, and smaller absolute values $|\mathrm{\gls{mcd}}|$ correspond to less severe breaches.
We use \gls{mcd+} to denote this signed performance interpretation, where smaller positive values are preferred as a proxy for minimal intervention, and negative values (especially large negative ones) are undesirable (hence the ``+'' sign).

\subsection{Research Questions}

We investigate the following research questions in this paper.
\begin{description}

\item[\textbf{RQ1.}] \textit{How do different \gls{piml} models and analytical models perform on the geofencing task?}
This question compares the performance of various physics-informed machine learning models and analytical models in maintaining containment within geofenced boundaries.

\vspace{0.5em}
\item[\textbf{RQ2.}] \textit{How do different operating regimes (speed and steering curvature) influence the performance of different models?}
This question examines how various models perform across four distinct regimes: high-speed--sharp-steering, high-speed--straight, low-speed--sharp-steering, and low-speed--straight.

\vspace{0.5em}

\item[\textbf{RQ3.}] \textit{How does the control-affine network structure affect control linearity and safety performance?}
This question investigates whether decoupling the residual learning of $f$ (drift dynamics) and $g$ (control influence) improves control linearity and containment performance compared to a conventional residual neural network.
Specifically, it compares architectures where $f$ and $g$ share the same network against those where they are modeled by separate subnetworks within the control-affine formulation.

\vspace{0.5em}

\item[\textbf{RQ4.}] \textit{How do key hyperparameters, including contraction rate ($\gamma$) and model size, influence the performance?}
This question explores how tuning $\gamma$ and scaling model size affect the performance metrics and whether optimal $\gamma$ values generalize across models and vehicles.

\end{description}

\begin{figure*}[!ht]
    \centering
    \begin{subfigure}[b]{0.8\textwidth}
        \centering
        \includegraphics[width=\textwidth]{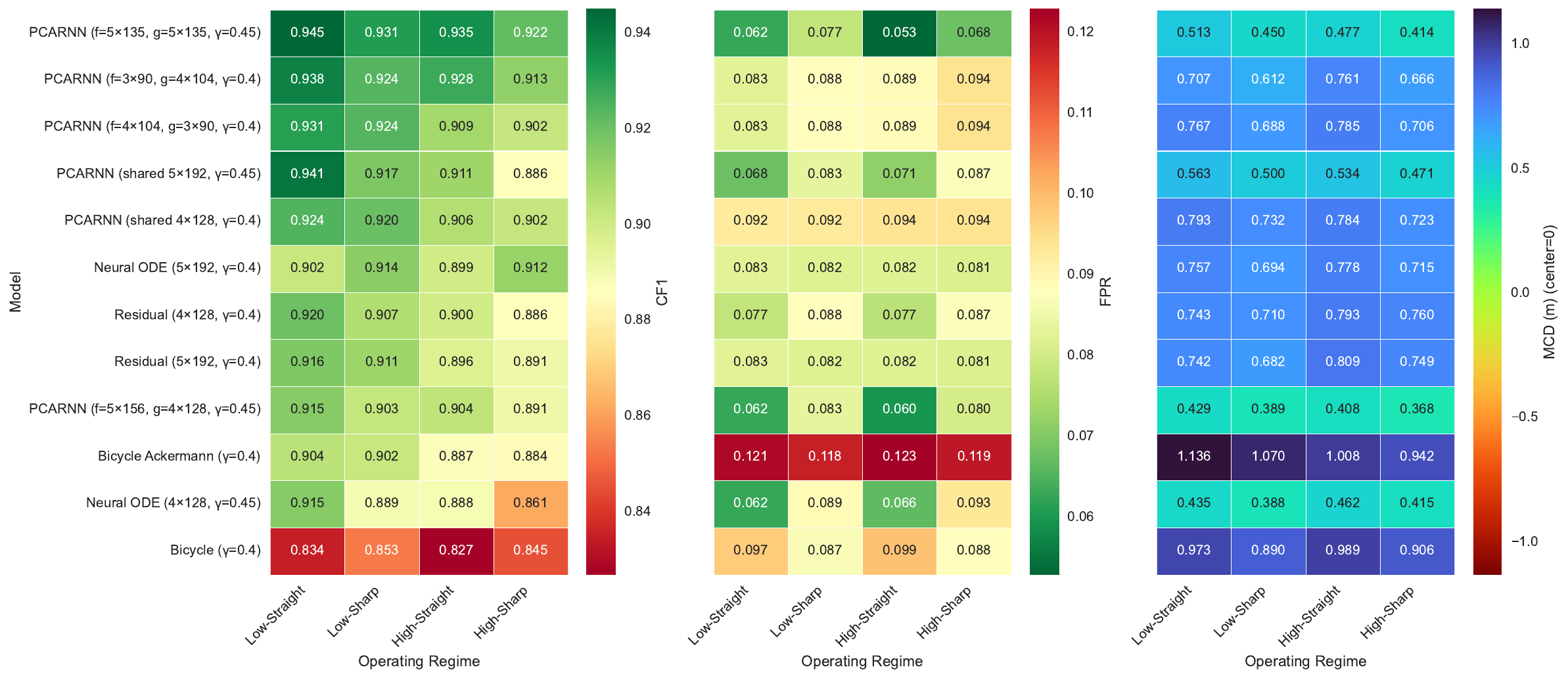}
        \caption{Stratified multi-metric analysis for the Audi E-tron across operating regimes.
        Each heatmap shows \gls{cf1}, \gls{fpr}, and \gls{mcd+} scores for different model architectures under varying actuation regimes.
        Higher \gls{cf1} and lower \gls{fpr}/\gls{mcd+} indicate better containment and smoother control responses.}
        \label{fig:stratified_audi}
    \end{subfigure}

    \vspace{1em}

    \begin{subfigure}[b]{0.8\textwidth}
        \centering
        \includegraphics[width=\textwidth]{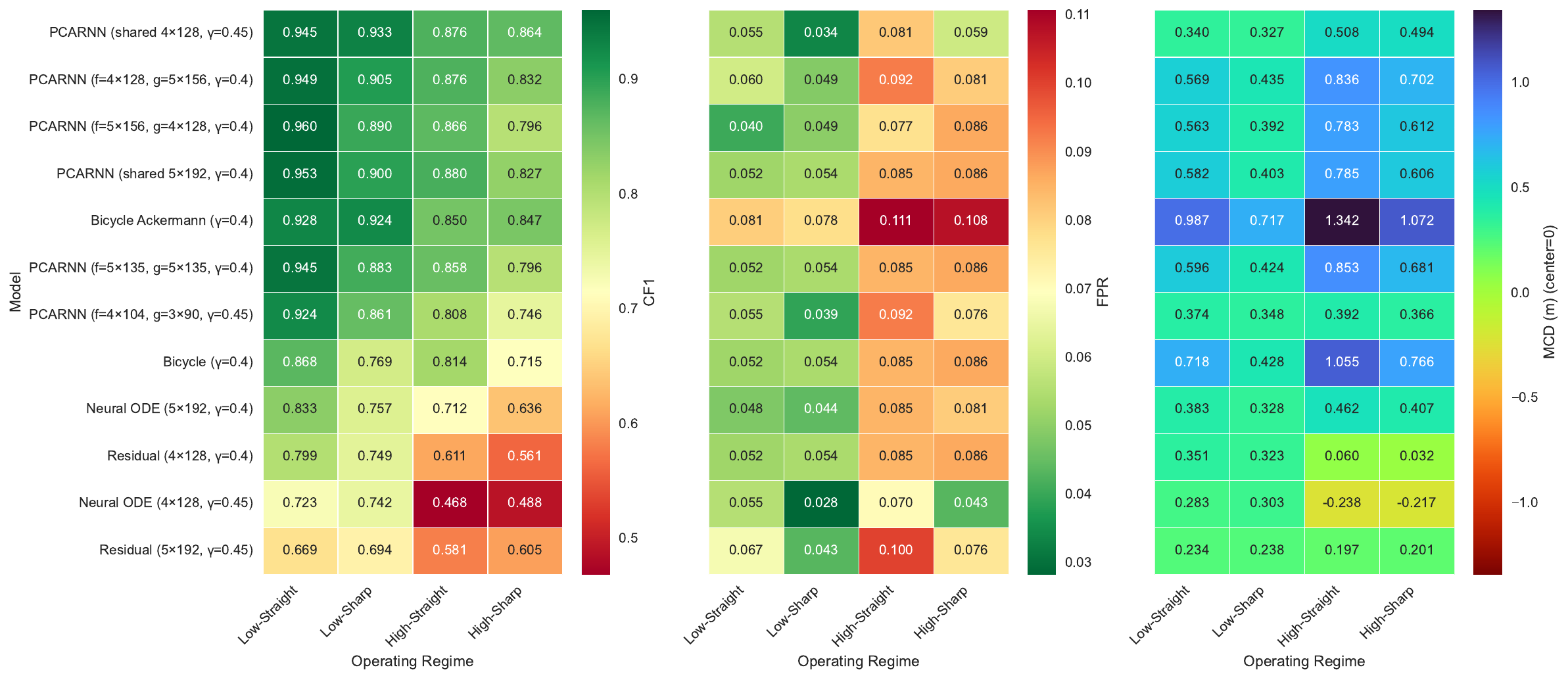}
        \caption{Stratified multi-metric analysis for the Lincoln MKZ across operating regimes.
        Results show the interaction between model architecture, actuation regime, and control linearity in determining containment performance.}
        \label{fig:stratified_lincoln}
    \end{subfigure}

    \caption{Comparison of containment performance, intervention rate, and control smoothness across operating regimes for the Audi E-tron and Lincoln MKZ.
    Each panel illustrates how different architectures and control-affine formulations perform under low/high speed and steering conditions.}
    \label{fig:stratified_multi_metric}
\end{figure*}
\subsection{Main Results (RQ1)}
Table~\ref{tab:regime_summary} summarizes performance across vehicles, hyperparameters, and model variants.
Overall, \gls{pcarnn} variants dominate on containment safety: they achieve the strongest \gls{cf1} on both vehicles while keeping \gls{fpr} near the best and \gls{mcd} competitive. This indicates that preserving the control–affine structure and learning residuals for both drift and control yields better calibrated decision boundaries than purely data-driven dynamics or \gls{piml} with combined residuals.

Second, the split vs. shared design interacts with vehicle dynamics. On the Audi E-tron, the split variant benefits from extra capacity, suggesting that independently tailoring drift and control residuals helps a vehicle with richer actuation–dynamics couplings. On the Lincoln MKZ, the smaller shared network peaks in \gls{cf1}, implying that parameter sharing can regularize learning when a simpler inductive bias matches the vehicle’s dynamics. In both cases, capacity is useful only to the extent that it respects the control–affine inductive bias; scaling without structure does not close the gap.

Third, alternative baselines expose a safety–conservatism trade-off. Residual models without control–affine constraints tend to be conservative, often improving \gls{fpr} and \gls{mcd} locally, but they compromise \gls{cf1}, pointing to under-detection of unsafe cases. Neural \gls{ode} loses separability between state and input, which harms overall safety classification despite expressive function approximation. Classical bicycle models trail in \gls{cf1} and are less consistent on \gls{mcd}, highlighting the limitation of fixed-form physics without learned corrections.

\noindent{\color{black}\textbf{\underline{Findings (RQ1)}}
Control–affine architectures (\gls{pcarnn}) consistently outperform both analytical and other \gls{piml} baselines in containment safety, achieving the best balance between \gls{cf1}, \gls{fpr}, and \gls{mcd}. 
Preserving separability between drift and control enables more stable and interpretable safety boundaries than residual or neural \gls{ode} models.
Model capacity and sharing strategy interact with vehicle dynamics: electric drivetrains (Audi) benefit from split $f$–$g$ specialization, while combustion drivetrains (Lincoln) favor shared representations that regularize the coupled propulsion–braking behavior.}

\subsection{Performance on Different Driving Scenarios (RQ2)}

The stratified analysis in Figure~\ref{fig:stratified_multi_metric} highlights how containment, intervention rate, and control smoothness vary across operating regimes and vehicle types.
For the {Audi E-tron}, split control-affine models with separate $f$ and $g$ components achieve the highest containment (\gls{cf1} $\approx$ 0.93--0.94) and lowest intervention rates, particularly under high-speed and sharp-turn scenarios.
This reflects the electric drivetrain’s rapid and linear torque response, where modeling $f$ and $g$ independently improves stability and precision.
For the {Lincoln MKZ}, shared-network architectures perform best overall, especially in sharp-turn conditions, suggesting that the coupled throttle and braking dynamics of petrol vehicles benefit from a jointly learned control-affine representation of drift and control effects.
Overall, these results confirm that \textit{the optimal model structure depends on the vehicle’s actuation characteristics and dynamic coupling}, demonstrating the adaptability of the control-affine formulation across heterogeneous platforms.

\noindent{\color{black}\textbf{\underline{Findings (RQ2)}}
Performance varies systematically across driving regimes. 
Split \gls{pcarnn} models excel on the Audi E-tron, particularly in high-speed or sharp-turn conditions, where independent modeling of $f$ and $g$ captures the vehicle’s fast, decoupled actuation.
The Lincoln MKZ achieves stronger containment with shared networks, especially in nonlinear low-speed regimes, reflecting tighter coupling between throttle and braking.}

\subsection{The Impact of the Control-Affine Structure (RQ3)}

As established in Sec.~\ref{sec:related}, a key challenge for discrete-time CBFs is that generic nonlinear dynamics lead to intractable optimizations \cite{agrawal_discrete_2017}. A known remedy is to enforce a control-affine structure, which recovers a tractable and well-behaved QP \cite{khajenejad_tractable_2021}.

The \gls{dcbf} controller (Eq.~\eqref{eqa:qp}) relies on this control-affine assumption: small variations in $u=[\dot{\delta},F_x]^\top$ should induce an approximately linear change in the terminal barrier $h$. If this does not hold, the optimization may use inaccurate sensitivities, leading to constraint violations or undue conservatism. In this section, we directly test our hypothesis: does the \gls{pcarnn} architecture, by design, produce a more linear control response (and thus better \gls{dcbf} compatibility) than a generic residual or neural ODE baseline?

\subsubsection{Gain of the control-affine structure with and without split $f$ and $g$ networks}
We conduct an ablation study to assess the contribution of the control-affine formulation to the geofencing task. Specifically, we compare a standard residual \gls{piml} network (without the control-affine structure) against variants that incorporate control-affine decomposition, where the drift dynamics $f$ and control influence $g$ are either represented by a shared network or by separate subnetworks.
Table~\ref{tab:control_affine_ablation} summarizes the results. Numbers in parentheses indicate performance changes relative to the residual baseline, with improvements highlighted in green and declines in red.

Across all tested configurations, introducing the control-affine structure markedly improved the containment safety metric (\gls{cf1}) compared to the residual baseline.
Among the control-affine variants, the split-network design achieved the highest overall performance, indicating that explicitly modeling the control influence $g$ as a linear component enhances both stability and responsiveness.
The split formulation offers greater flexibility by allowing the drift dynamics $f$ and control influence $g$ to specialize in distinct behaviors: $f$ can better capture the nonlinear passive dynamics of the vehicle, while $g$ focuses on the control-dependent response, improving adaptation under varying actuation modes (e.g. braking-only vs.\ braking+steering).
This separation reduces interference between the learned dynamics and control pathways, resulting in smoother barrier updates and more consistent containment across vehicles.
Overall, the results confirm that the control-affine formulation provides a substantial performance gain for geofencing applications without increasing model complexity.

\begin{table*}[t]
\centering
\caption{\label{tab:control_affine_ablation}%
Ablation study on the effect of control-affine components. The residual \gls{piml} network serves as the baseline. Variants include control-affine formulations where the drift dynamics $f$ and control influence $g$ are either modeled by a shared network or by separate subnetworks of comparable parameter count. Numbers in parentheses denote changes relative to the baseline; green indicates improvement and red indicates decline.}
\resizebox{\textwidth}{!}{
\begin{tabular}{
    p{4cm}
    p{2.5cm}
    p{0.8cm}
    c
    c
    c c
    c c
}
\toprule
& & & \multicolumn{2}{c}{\textbf{Braking + Steering}} & \multicolumn{4}{c}{\textbf{Braking}} \\
\cmidrule(lr){4-5} \cmidrule(lr){6-9}
\textbf{Variant} & \textbf{Size} & \textbf{Params} &
\textbf{Lincoln MKZ} &
\textbf{Audi E-tron} &
\multicolumn{2}{c}{\textbf{Lincoln MKZ}} &
\multicolumn{2}{c}{\textbf{Audi E-tron}} \\
\cmidrule(lr){4-4} \cmidrule(lr){5-5} \cmidrule(lr){6-7} \cmidrule(lr){8-9}
& & &
CBF &
CBF &
{TTC} & {CBF} &
{TTC} & {CBF} \\
\midrule
Residual & $5{\times}192$ & 150K & 0.748  & 0.913 & \underline{0.623} & 0.624 & 0.449 & 0.571 \\
~ w/ Control-affine (shared network)  & $5{\times}192$ & 150K &
{\textbf{0.913}} {\color{green!50!black}(+0.165)} &
0.918 {\color{green!50!black}(+0.005)} &
\textbf{0.806} {\color{green!50!black}(+0.183)} &
\underline{0.634} {\color{green!50!black}(+0.010)} &
\underline{0.802} {\color{green!50!black}(+0.353)} &
0.595 {\color{green!50!black}(+0.024)} \\
~ w/ Control-affine (split networks)  & f=$5{\times}135$, g=$5{\times}135$ & 149K &
{0.873} {\color{green!50!black}(+0.125)} &
\textbf{0.935} {\color{green!50!black}(+0.022)} &
0.562 {\color{red!}(-0.061)} &
\underline{0.634} {\color{green!50!black}(+0.010)} &
0.794 {\color{green!50!black}(+0.345)} &
\underline{0.603} {\color{green!50!black}(+0.032)} \\
~ w/ Control-affine (split networks) & f=$4{\times}128$, g=$5{\times}156$ & 150K &
\textbf{0.913} {\color{green!50!black}(+0.165)} &
\underline{0.919} {\color{green!50!black}(+0.006)} &
{0.536} {\color{red!}(-0.087)} &
\textbf{0.640} {\color{green!50!black}(+0.016)} &
\textbf{0.803} {\color{green!50!black}(+0.354)} &
\textbf{0.612} {\color{green!50!black}(+0.041)} \\
~ w/ Control-affine (split networks) & f=$5{\times}156$, g=$4{\times}128$ & 150K &
\underline{0.907} {\color{green!50!black}(+0.159)} &
\underline{0.919} {\color{green!50!black}(+0.006)} &
0.552 {\color{red!}(-0.071)} &
\textbf{0.640} {\color{green!50!black}(+0.016)} &
0.798 {\color{green!50!black}(+0.349)} &
{0.609} {\color{green!50!black}(+0.038)} \\
\bottomrule
\end{tabular}
}
\end{table*}

\begin{figure*}[t]
    \centering
        \begin{subfigure}[b]{0.48\textwidth}
        \centering
        \includegraphics[width=\linewidth]{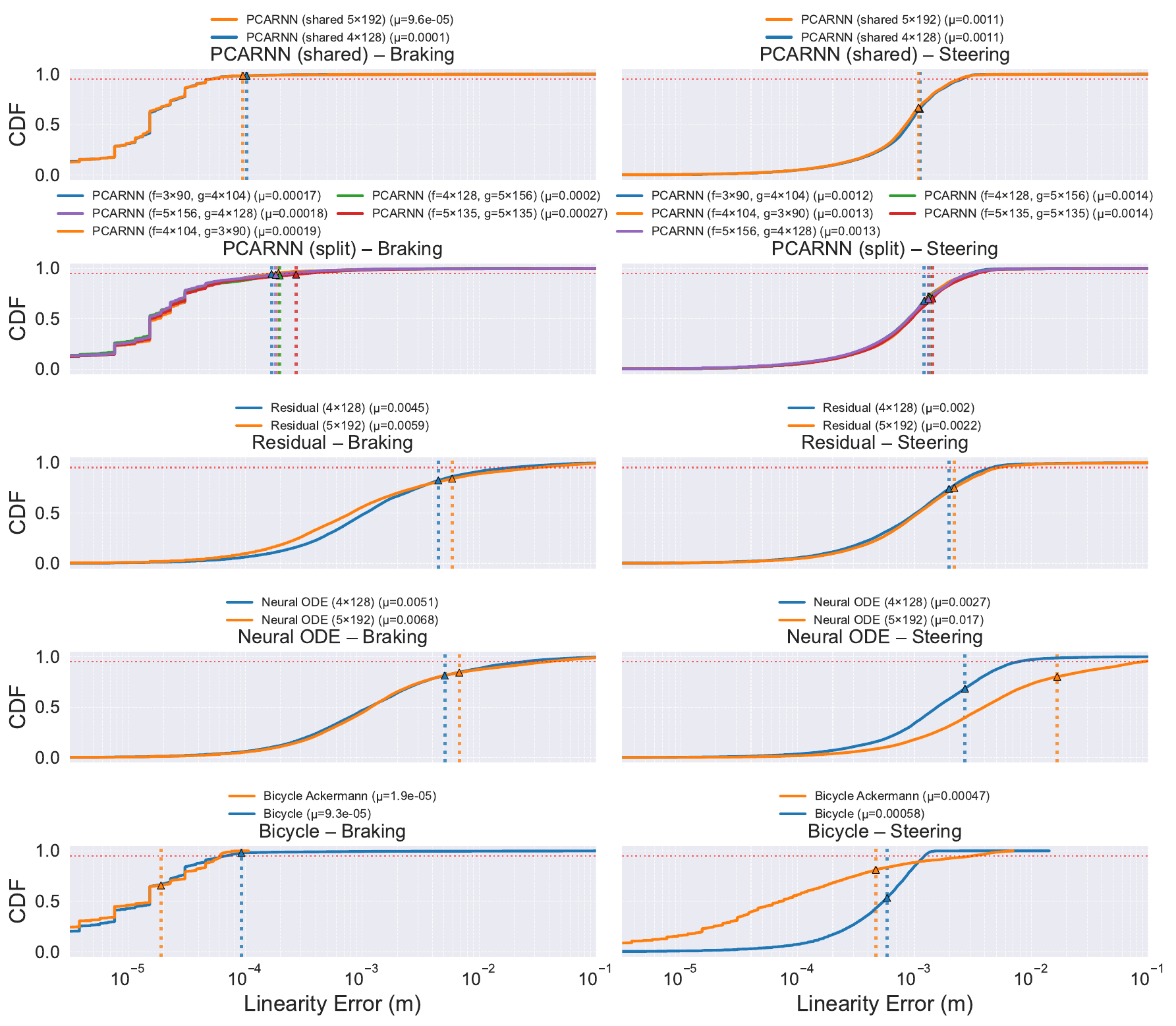}
        \caption{CDF of control linearity $\epsilon_{\text{lin}}$ for the vehicle Audi E-tron.
        The electric drivetrain exhibits tighter control-affine behavior, particularly under higher $\gamma$.}
        \label{fig:linearity_audi}
    \end{subfigure}
    \hfill
    \begin{subfigure}[b]{0.48\textwidth}
        \centering
        \includegraphics[width=\linewidth]{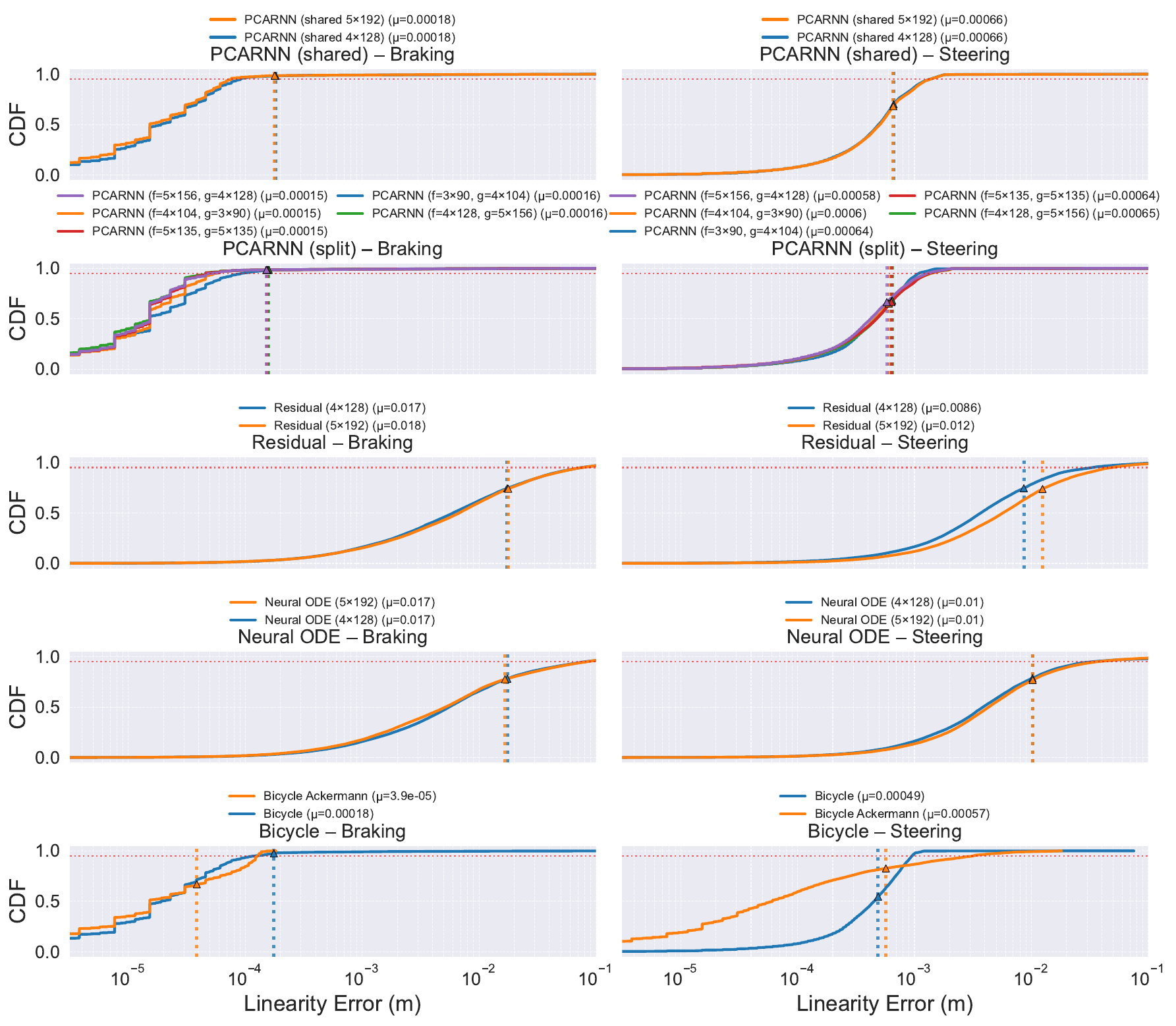}
        \caption{CDF of control linearity $\epsilon_{\text{lin}}$ for the vehicle Lincoln MKZ.
        Lower values indicate better local linearity and more stable control responses.}
        \label{fig:linearity_lincoln}
    \end{subfigure}
    \caption{Comparison of control linearity distributions for the Lincoln MKZ and Audi E-tron under varying $\gamma$ and model configurations.
    Each CDF shows how well the discrete control barrier function (\gls{dcbf}) maintains local linearity during operation.}
    \label{fig:linearity_cdf}
\end{figure*}

\begin{table}[ht]
\centering
\caption{\label{tab:alpha}Ablation on the hyperparameter $\gamma$ for the \gls{pcarnn} (split) + \gls{dcbf} model.}
\resizebox{0.5\textwidth}{!}{
\begin{tabular}{
  p{1.5cm}   
  p{1.2cm} 
  p{0.5cm} 
  p{0.6cm}   
  p{0.6cm}   
  p{0.8cm}   
}
\toprule
\textbf{Size} & \textbf{Vehicle} 
& $\mathbf{\gamma}$ & \textbf{CF1}${\uparrow}$ & \textbf{FPR}${\downarrow}$ & \textbf{MCD}$^{+}{\downarrow}$ \\
\midrule
\multirow{6}{*}{\makecell{\(f=5\times135\)\\  \(g=5\times135\)}}
& \multirow{6}{*}{Audi E-tron} 
& 0.40 & 0.911 & 0.087 & 0.697 \\
&  & 0.44 & \underline{0.922} & 0.071 & 0.496 \\
  & & 0.45 & \textbf{0.935} & 0.066 & 0.470 \\
 & & 0.46 & 0.901 & 0.077 & 0.416 \\
 & & 0.48 & 0.896 & \underline{0.060} & \underline{0.355} \\
 & & 0.60 & 0.803 & \textbf{0.055} & \textbf{0.122} \\
\bottomrule
\end{tabular}
}
\end{table}

\begin{table*}[ht]
\centering
\caption{\label{tab:split_sizes}Ablation on \gls{pcarnn} model architecture design with the \gls{dcbf} controller. For each (Size $f$, Size $g$) configuration, only the $\gamma$ yielding the highest \gls{cf1} is shown (breaking ties with lower \gls{fpr} and \gls{mcd+}). Lower is better for \gls{fpr} and \gls{mcd+}; higher is better for  \gls{cf1}. Values in parentheses indicate improvement relative to the shared baseline at the same $\gamma$.}
\resizebox{1\textwidth}{!}{
\begin{tabular}{
  p{2cm}
  p{2.4cm}
  p{1.2cm}
  p{2cm}
  p{2cm}
  p{1cm}
  p{1.8cm}
  p{1.8cm}
  p{1.8cm}
}
\toprule
\textbf{Vehicle} & \textbf{Model} & \textbf{\gls{dcbf} $\gamma$} & \textbf{Size $f$} & \textbf{Size $g$} & \textbf{Params} & \textbf{CF1}${\uparrow}$ & \textbf{FPR}${\downarrow}$ & \textbf{MCD}$^{+}{\downarrow}$ \\
\midrule

\multirow{7}{*}{Audi E-tron}
& PCARNN (shared) & 0.40 & \multicolumn{2}{c}{$4\times128$} & 51.3K & 0.915 & \textbf{0.071} & 0.758 \\
& PCARNN (split) & 0.40 & \(3\times90\) & \(4\times104\) & 51K & \textbf{0.928} {\color{green!50!black}(+0.013)} & 0.087 {\color{red}(+0.016)} & 0.672 {\color{green!50!black}(-0.086)} \\
& PCARNN (split) & 0.40 & \(4\times104\) & \(3\times90\) & 51K & 0.921 {\color{green!50!black}(+0.006)} & 0.087 {\color{red}(+0.016)} & \textbf{0.612} {\color{green!50!black}(-0.146)} \\
\cline{2-9}
\addlinespace[4pt]
& PCARNN (shared) & 0.45 & \multicolumn{2}{c}{$5\times192$} & 150K & 0.918 & 0.077 & 0.498 \\
& PCARNN (split) & 0.45 & \(5\times135\) & \(5\times135\) & 149K & \textbf{0.935} {\color{green!50!black}(+0.017)} & \textbf{0.066} {\color{green!50!black}(-0.011)} & 0.470 {\color{green!50!black}(-0.028)} \\
& PCARNN (split) & 0.40 & \(4\times128\) & \(5\times156\) & 150K & 0.919 {\color{green!50!black}(+0.004)} & 0.087 {\color{red}(+0.016)} & {0.658} {\color{red}(+0.160)} \\
& PCARNN (split) & 0.45 & \(5\times156\) & \(4\times128\) & 150K & 0.905 {\color{red}(-0.013)} & 0.071 {\color{green!50!black}(-0.006)} & \textbf{0.396} {\color{green!50!black}(-0.102)} \\
\midrule

\multirow{7}{*}{Lincoln MKZ}
& PCARNN (shared) & 0.45 & \multicolumn{2}{c}{$4\times128$} & 51.3K & \textbf{0.922} & \textbf{0.050} & \textbf{0.356} \\
& PCARNN (split) & 0.45 & \(3\times90\) & \(4\times104\) & 51K & 0.908 {\color{red}(-0.014)} & 0.069 {\color{red}(+0.019)} & 0.647 {\color{red}(+0.291)} \\
& PCARNN (split) & 0.45 & \(4\times104\) & \(3\times90\) & 51K & 0.903 {\color{red}(-0.019)} & {0.062} {\color{red}(+0.012)} & {0.536} {\color{red}(+0.180)} \\
\cline{2-9}
\addlinespace[4pt]
& PCARNN (shared) & 0.40 & \multicolumn{2}{c}{$5\times192$} & 150K & \textbf{0.913} & 0.062 & 0.569 \\
& PCARNN (split) & 0.40 & \(5\times135\) & \(5\times135\) & 149K & 0.898 {\color{red}(-0.015)} & 0.062 {\color{black!}(+0.000)} & 0.552 {\color{green!50!black}(-0.017)} \\
& PCARNN (split) & 0.40 & \(4\times128\) & \(5\times156\) & 150K & \textbf{0.913} {\color{black!}(+0.000)} & 0.062 {\color{black!}(+0.000)} & {0.535} {\color{green!50!black}(-0.034)} \\
& PCARNN (split) & 0.40 & \(5\times156\) & \(4\times128\) & 150K & 0.882 {\color{red}(-0.031)} & \textbf{0.050} {\color{green!50!black}(-0.012)} & \textbf{0.318} {\color{green!50!black}(-0.251)} \\
\bottomrule
\end{tabular}}
\end{table*}

\subsubsection{Control linearity analysis for \gls{dcbf} compatibility}

To test to what extent that each learned dynamics model satisfies the linearity assumption, we perform a \textit{control linearity analysis}. This test evaluates whether the terminal barrier value $h$ changes approximately linearly in response to small control perturbations. Specifically, we measure how well the first-order approximation
\[
\Delta h \approx J \cdot \Delta u,
\]
holds, where $J = \nabla_u h$ is the Jacobian of $h$ with respect to the control inputs $u$.

For each state in the test dataset, we estimate $J$ numerically via central differences about the nominal control $u_{\text{nom}}$. The perturbations are applied over a $0.3\,\text{s}$ preview horizon with $\Delta \dot{\delta} = \pm 0.25\,\text{rad/s}$ and $\Delta F_x = \pm 800\,\text{N}$. We then propagate the full nonlinear dynamics under both $u_{\text{nom}}$ and $u_{\text{nom}} + \Delta u$ to obtain the actual change:
\[
\Delta h_{\text{actual}} = h(\Phi(s_0, u_{\text{nom}} + \Delta u)) - h(\Phi(s_0, u_{\text{nom}})),
\]
where $\Phi$ denotes the system rollout. The predicted change is given by
\[
\Delta h_{\text{predicted}} = J \cdot \Delta u,
\]
and the deviation between the two defines the \textit{linearity error}:
\[
\epsilon_{\text{lin}} = \lvert \Delta h_{\text{actual}} - \Delta h_{\text{predicted}} \rvert.
\]

Lower values of $\epsilon_{\text{lin}}$ indicate that the model exhibits a stable, approximately linear response to control perturbations, aligning with the \gls{dcbf}’s control-affine assumption. Models with high $\epsilon_{\text{lin}}$, on the other hand, violate this assumption and may produce unreliable or unstable barrier updates when deployed with the \gls{dcbf} controller.

Figure~\ref{fig:linearity_cdf} show the cumulative distributions of linearity error $\epsilon_{\text{lin}}$ under braking and steering perturbations.
Lower $\epsilon_{\text{lin}}$ values indicate a more linear, control-affine response, aligning better with the \gls{dcbf} assumption.

Across both vehicles, braking and steering responses behave differently. 
\textbf{Braking} perturbations exhibit larger dispersion in $\epsilon_{\text{lin}}$, especially for learned models, reflecting the stronger nonlinear coupling between longitudinal force and vehicle dynamics.
\textbf{Steering} perturbations, by contrast, show much tighter CDFs with mean errors often one to two orders of magnitude smaller, indicating that lateral dynamics are locally smoother and more linear.

For the \textit{Audi E-tron} (cf. Figure~\ref{fig:linearity_audi}), both shared and split \gls{pcarnn} variants achieve the lowest linearity errors across braking and steering, maintaining $\mu \approx 10^{-4}$--$10^{-3}$\,m.
The split architecture provides slightly tighter consistency, suggesting that explicitly modeling $f$ and $g$ separately helps capture the EV’s decoupled torque and steering dynamics.
The analytical bicycle model performs comparably well, confirming that the control-affine assumption remains valid for this drivetrain.
In contrast, residual and neural \gls{ode} models display broader error distributions—especially under braking—indicating higher nonlinearity and reduced \gls{dcbf} compatibility.

For the \textit{Lincoln MKZ} (cf. Figure~\ref{fig:linearity_lincoln}), overall linearity errors are higher, particularly under braking, reflecting the more complex drivetrain dynamics of the combustion vehicle.
Here, the shared \gls{pcarnn} variant achieves the best linearity performance ($\mu \approx 10^{-5}$–$10^{-4}$\,m), while the split version shows modest degradation.
This suggests that jointly learning $f$ and $g$ stabilizes the interaction between throttle and brake channels, which are inherently coupled in the petrol configuration.
Residual and neural \gls{ode} models again exhibit the largest variability, confirming their limited ability to maintain consistent linear response under nonlinear longitudinal coupling.

\noindent
{\color{black}{\underline{\textbf{Findings (RQ3)}}}
Braking introduces greater nonlinearity than steering across all models.
Control-affine formulations (analytical and \gls{pcarnn}) maintain low and consistent $\epsilon_{\text{lin}}$, validating their suitability for \gls{dcbf}-based control.
Electric drivetrains (Audi) benefit from separated $f$–$g$ learning, as their torque response is fast and largely decoupled from steering, enabling each subnetwork to specialize in distinct control effects. 
Combustion drivetrains (Lincoln) achieve better stability with shared architectures, since throttle, braking, and load transfer are inherently coupled and require joint representation of drift and control dynamics.}

\subsection{Contraction Rate $\gamma$ and Model Size (RQ4)}

The sweep over contraction rates $\gamma$ and model sizes highlights consistent trade-offs between safety enforcement, responsiveness, and control smoothness.
Values around $\gamma=0.45$ deliver the best overall containment performance for both vehicles, achieving strong safety enforcement without excessive conservatism.
Lower $\gamma$ values make the controller more permissive, sometimes allowing delayed reactions, while higher values ($\gamma>0.5$) produce over-constrained, abrupt interventions that reduce responsiveness and increase containment oscillations.

Model capacity also affects performance stability.
Compact networks ($\sim$50K parameters) perform competitively, benefiting from the structured control-affine design that efficiently separates drift and control effects.
Larger networks ($\sim$150K parameters) improve containment (\gls{cf1}) and reduce false positives (\gls{fpr}), particularly when paired with well-tuned $\gamma$ values, but can exhibit mild over-sensitivity under high contraction rates.

\noindent{\color{black}{\textbf{\underline{Findings (RQ4)}} Optimal contraction rates cluster around $\gamma\approx0.45$ across models and vehicles, representing a robust equilibrium between safety and responsiveness. Moderate network sizes provide the most stable and generalizable performance. }} 
\begin{table*}[ht]
\centering
\caption{\label{tab:hyperparameter}Ablation on model architecture variants with the \gls{dcbf} controller. Lower is better for \gls{fpr} and \gls{mcd}, and higher is better for \gls{cf1}.}
\resizebox{1\textwidth}{!}{
\begin{tabular}{
  p{1.8cm}   
  p{4cm}   
  p{4cm}   
  p{1.5cm}   
  p{1.5cm}   
  p{1.5cm}   
  p{1.5cm}   
}
\toprule
\textbf{Vehicle} & \textbf{Variant} & \textbf{Size} & \textbf{Params} & \textbf{CF1} ${\uparrow}$  & \textbf{FPR} ${\downarrow}$  & \textbf{MCD}$^{+}{\downarrow}$ \\
\midrule

\multirow{23}{*}{Audi E-tron}
& PCARNN (shared) (\(\gamma=0.40\)) & \(4\times 128\)          & 51.3K & 0.915 & 0.071 & 0.758 \\
& PCARNN (shared) (\(\gamma=0.40\)) & \(5\times 192\)          & 150K  & 0.907 & 0.093 & 0.798 \\
& PCARNN (shared) (\(\gamma=0.45\)) & \(4\times 128\)          & 51.3K & 0.915 & 0.082 & 0.438 \\
& PCARNN (shared) (\(\gamma=0.45\)) & \(5\times 192\)          & 150K & 0.918 & 0.077 & 0.498 \\
  \cline{2-7}
\addlinespace[4pt]
& PCARNN (split) (\(\gamma=0.40\)) & \(f=3\times 90,\; g=4\times 104\) & 51K & \underline{0.928} & 0.087 & 0.672 \\

& PCARNN  (split) (\(\gamma=0.40\)) & \(f=5\times 135,\; g=5\times 135\) & 149K & 0.911 & 0.087 & 0.697 \\

& PCARNN  (split) (\(\gamma=0.40\)) & \(f=4\times 128,\; g=5\times 156\) & 150K & 0.919 & 0.087 & 0.658 \\
& PCARNN  (split) (\(\gamma=0.45\)) & \(f=3\times 90,\; g=4\times 104\)  & 51K & 0.915 & 0.071 & 0.458 \\
& PCARNN  (split) (\(\gamma=0.45\)) & \(f=5\times 135,\; g=5\times 135\) & 149K & \textbf{0.935} & \underline{0.066} & 0.470 \\
& PCARNN  (split) (\(\gamma=0.45\)) & \(f=4\times 128,\; g=5\times 156\) & 150K & 0.914 & 0.071 & \textbf{0.410} \\

  \cline{2-7}
\addlinespace[4pt]

& Residual (\(\gamma=0.40\)) & \(4\times 128\)  & 50.9K & 0.906 & 0.082 & 0.706 \\
& Residual (\(\gamma=0.40\)) & \(5\times 192\)  & 150K & 0.906 & 0.082 & 0.731 \\
& Residual (\(\gamma=0.45\)) & \(4\times 128\)  & 50.9K & 0.892 & \underline{0.066} & \underline{0.412} \\
& Residual (\(\gamma=0.45\)) & \(5\times 192\)  & 150K & 0.913 & \textbf{0.060} & 0.489 \\
  \cline{2-7}
\addlinespace[4pt]

& Neural ODE (\(\gamma=0.40\)) & \(4\times 128\) & 50.9K & 0.910 & 0.077 & 0.734 \\
& Neural ODE (\(\gamma=0.40\)) & \(5\times 192\) & 150K & 0.906 & 0.082 & 0.746 \\

& Neural ODE (\(\gamma=0.45\)) & \(4\times 128\) & 50.9K & 0.893 & 0.077 & 0.425 \\
& Neural ODE (\(\gamma=0.45\)) & \(5\times 192\) & 150K & 0.887 & \underline{0.066} & 0.617 \\
\cline{2-7}
  \addlinespace[4pt]

& Bicycle (\(\gamma=0.40\)) & N/A & N/A & 0.839 & 0.093 & 0.964 \\
& Bicycle (\(\gamma=0.45\)) & N/A & N/A & 0.850 & 0.077 & 0.617 \\
& Bicycle Ackermann (\(\gamma=0.40\)) & N/A & N/A & 0.896 & 0.120 & 1.080 \\
\toprule
  \multirow{23}{*}{Lincoln MKZ}
& PCARNN (shared) (\(\gamma=0.40\)) & \(4\times 128\) & 51.3K & \underline{0.913} & 0.062 & 0.558 \\
& PCARNN (shared) (\(\gamma=0.40\)) & \(5\times 192\) & 150K & \underline{0.913} & 0.062 & 0.569 \\
& PCARNN (shared) (\(\gamma=0.45\)) & \(4\times 128\) & 51.3K & \textbf{0.922} & \underline{0.050} & 0.356 \\
& PCARNN (shared) (\(\gamma=0.45\)) & \(5\times 192\) & 150K & 0.896 & 0.056 & 0.502 \\\cline{2-7}
\addlinespace[4pt]
& PCARNN  (split) (\(\gamma=0.40\)) & \(f=3\times 90,\; g=4\times 104\) & 51K & {0.908} & 0.069 & 0.647 \\
& PCARNN  (split) (\(\gamma=0.40\)) & \(f=5\times 135,\; g=5\times 135\) & 149K & 0.898 & 0.062 & 0.552 \\
& PCARNN  (split) (\(\gamma=0.40\)) & \(f=4\times 128,\; g=5\times 156\) & 150K & \underline{0.913} & 0.062 & 0.535 \\
& PCARNN  (split) (\(\gamma=0.45\)) & \(f=3\times 90,\; g=4\times 104\) & 51K & 0.906 & 0.056 & 0.534 \\
& PCARNN  (split) (\(\gamma=0.45\)) & \(f=5\times 135,\; g=5\times 135\) & 149K & 0.873 & \underline{0.050} & 0.343 \\
& PCARNN  (split) (\(\gamma=0.45\)) & \(f=4\times 128,\; g=5\times 156\) & 150K & 0.892 & \underline{0.050} & 0.340 \\

  \cline{2-7}
\addlinespace[4pt]
& Residual (\(\gamma=0.40\)) & \(4\times 128\) & 50.9K & 0.728 & 0.062 & 0.309 \\
& Residual (\(\gamma=0.40\)) & \(5\times 192\) & 150K & 0.748 & 0.062 & 0.341 \\
& Residual (\(\gamma=0.45\)) & \(4\times 128\) & 50.9K & 0.737 & \textbf{0.044} & 0.293 \\
& Residual (\(\gamma=0.45\)) & \(5\times 192\) & 150K & 0.653 & 0.062 & \textbf{0.235} \\\cline{2-7}
\addlinespace[4pt]
& Neural ODE (\(\gamma=0.40\)) & \(4\times 128\) & 50.9K & 0.718 & 0.062 & 0.344 \\
& Neural ODE (\(\gamma=0.40\)) & \(5\times 192\) & 150K & 0.771 & 0.056 & 0.359 \\
& Neural ODE (\(\gamma=0.45\)) & \(4\times 128\) & 50.9K & 0.660 & \textbf{0.044} & \underline{0.253} \\
& Neural ODE (\(\gamma=0.45\)) & \(5\times 192\) & 150K & 0.716 & \underline{0.050} & 0.284 \\\cline{2-7}
\addlinespace[4pt]
& Bicycle (\(\gamma=0.40\)) & N/A & N/A & 0.816 & 0.062 & 0.684 \\
& Bicycle (\(\gamma=0.45\)) & N/A & N/A & 0.832 & \textbf{0.044} & 0.457 \\
& Bicycle Ackermann (\(\gamma=0.40\)) & N/A & N/A & 0.905 & 0.088 & 0.999 \\

\bottomrule
\end{tabular}}
\end{table*}

\section{Conclusion}
We presented \gls{pcarnn-dcbf}, a geofencing framework that integrates a hybrid control–affine dynamics model with a preview-based safety filter to enforce polygonal keep–in constraints. By explicitly preserving the control–affine structure within the learning loop, our pipeline ensures that the safety constraint remains linear in the control input. This structural prior enables the use of a fast, interpretable \gls{qp} for real-time intervention, avoiding the tractability issues common to unstructured neural models while correcting the fidelity gaps of purely analytical baselines.

Our comparative analysis highlighted the trade-offs inherent in data-driven safety filtering. While Neural \glspl{ode} offer high flexibility, they obscure the direct influence of actuation, leading to inconsistent safety margins. The PCARNN approach proved superior by maintaining the certifiable backbone of the bicycle model while learning targeted residuals for tire and load–transfer effects. This "structure-preserving" strategy yielded the highest containment reliability (\gls{cf1}) with modest data requirements.

Furthermore, we found that drivetrain topology dictates the optimal learning architecture. Decoupled electric powertrains favored split drift/control networks to maximize control authority, whereas coupled combustion dynamics required shared representations to regularize the interaction between propulsion and braking. 

Despite the observed gains, these findings are subject to several limitations. We target static keep–in sets and do not model moving agents or social driving rules. The dynamics are a bicycle style abstraction validated in simulation; real vehicle effects from tires, load transfer, and low friction can be stronger. The preview length and exponential target in the \gls{dcbf} are engineered choices rather than formal recursive feasibility proofs.

Future work will extend the framework to dynamic environments with moving obstacles and incorporate robust state estimation to handle localization noise. As a direct next step, we aim to validate the full stack on a physical vehicle, specifically targeting low-friction winter conditions to test the generalization limits of the learned residuals.

\section*{Acknowledgment}

We thank Sweden’s Innovation Agency, Vinnova, for funding Grant No. 2021-05052, making the herein-described work possible.

\bibliographystyle{IEEEtran}
\bibliography{bibliography}
\appendix

\section{Technical Appendix: \gls{carla} Integration and Dataset Generation}
\label{sec:appendix_carla}

\subsection{Data Fidelity and Simulation Environment}

The experimental framework employs the \gls{carla} simulator (version 0.9.15)~\cite{dosovitskiy2017carla} with a custom Python interface to generate high-fidelity, kinematically consistent datasets for training the \gls{pcarnn}. This predictive model forms the core of the safety framework.

\subsubsection{\gls{carla} Physics Customization}

To ensure the simulator’s vehicle dynamics align with the analytical bicycle model assumptions and to avoid simulator-specific artifacts, \gls{carla}’s physics configuration was extensively customized:

\begin{itemize}
    \item \textbf{Actuator authority:} The default speed-dependent steering attenuation was flattened to allow full steering range ($\delta_{\text{max}}$) at all speeds. This was implemented by setting the \texttt{steering\_curve} field of the \texttt{carla.VehiclePhysicsControl} object to a constant value, replacing the default speed-dependent curve (e.g. \texttt{[[0.0, 1.0], [10.0, 0.5]]}) documented in the CARLA physics API~\cite{carla_python_api_physics}. This ensures that the controller maintains unrestricted authority.
    \item \textbf{Powertrain characterization:}
    The vehicle physics were parameterized to provide a consistent mapping between the controller’s commanded longitudinal force ($F_x$) and \gls{carla}’s throttle/brake inputs.
    For the \textit{Audi e-tron}, we used the default \textsc{carla} vehicle blueprint (\texttt{vehicle.audi.etron}) and its stock EV powertrain parameters~\cite{dosovitskiy2017carla,carla_vehicle_blueprints}.
    For the \textit{Lincoln MKZ (2017)}, we used powertrain and transmission parameters from the OpenCOOD SmartData Model~\cite{opencood_smartdata,xu2022opv2v}.
    \item \textbf{Fidelity alignment:} Tire friction coefficients, brake torque limits, and stiffness parameters were tuned so that the handling limits (maximum lateral acceleration and deceleration) matched the analytical assumptions used in the Discrete Control Barrier Function (\gls{dcbf}) formulation.
    The Audi E-tron retained \gls{carla}’s default chassis and tire setup, while the Lincoln MKZ used the OPV2V-derived physical parameters.
\end{itemize}

\subsubsection{Kinematically Consistent State and Derivative Extraction}

Rather than relying on \gls{carla}’s instantaneous acceleration outputs (which can exhibit numerical noise and drift), all ground-truth derivatives were computed via finite differencing to ensure consistency between $\mathbf{x}$, $\dot{\mathbf{x}}$, and $\mathbf{u}$:

\begin{itemize}
    \item \textbf{State definition:}
    The system state $\mathbf{x}$ includes position $(p_x, p_y)$, yaw $\psi$, body-frame velocities $(v_x, v_y)$, yaw rate $\omega$, and effective front wheel steering angle $\delta$.
    The effective $\delta$ is computed using a standard Ackermann steering relation from the individual left/right wheel angles to provide a kinematically equivalent bicycle model input~\cite{mathworks_steeringsystem}.
    \item \textbf{Derivative calculation:}
    State derivatives $\dot{\mathbf{x}} = [\dot{v}_x, \dot{v}_y, \dot{\omega}, \dot{\delta}]^{\top}$ are obtained using a three-point central or forward finite-difference scheme over a fixed time step $\Delta t = 0.02\,\text{s}$.
    This ensures that $\mathbf{x}_k$, $\mathbf{u}_k$, and $\dot{\mathbf{x}}_k$ are mathematically coupled, producing a consistent training signal for residual and ODE-based dynamics learning.
\end{itemize}

\subsection{Dataset Generation and Stratified Sampling}

The training dataset is generated through a \textbf{Robust Systematic Scenario Generator} that samples vehicle states and control sequences across a wide dynamic range, including near-tire-saturation regimes relevant to extrapolation testing.

\subsubsection{Systematic Scenario Space}

Three control dimensions were discretely sampled to ensure comprehensive coverage of operating conditions:

\begin{enumerate}
    \item \textbf{Initial longitudinal speed ($v_x$):} Six representative speeds ranging from $0.00$\,m/s to $35.00$\,m/s.
    \item \textbf{Steering input ($\dot{\delta}$):} Time-varying profiles including ramp steer (47.9\%), sinusoidal steer (41.7\%), constant-rate turns, and step inputs.
    \item \textbf{Longitudinal force ($F_x$):} Force commands covering step (23.3\%), constant (21.7\%), ramp (19.8\%), sinusoidal (18.1\%), and multi-phase sequences (17.1\%) simulating mixed braking and acceleration maneuvers.
\end{enumerate}

\subsubsection{Dataset Structuring and Filtering}

\begin{itemize}
    \item \textbf{Data size:}
    The final dataset includes $420$ unique scenarios, yielding approximately $1.25$ million data points for dense coverage of the vehicle’s dynamic envelope.
    \item \textbf{Inversion for symmetry:}
    Each scenario was mirrored laterally (sign inversion for lateral states and control inputs) to double the effective dataset and enforce left–right steering symmetry in the learned dynamics.
    \item \textbf{Realism filter:}
    Scenarios requiring non-physical inputs (e.g. excessively high-frequency steering at low speed or instantaneous stops) were excluded.
    The resulting dataset enforces physically realistic control limits, with maximum braking $F_x \approx -11{,}979$\,N and maximum acceleration $F_x \approx 7{,}000$\,N.
\end{itemize}

\subsubsection{Stratified Splitting for Generalization}

To ensure representative coverage across different motion patterns, the dataset is divided using stratified sampling:
\[
\text{Train:Val:Test} \approx 75\% : 12.5\% : 12.5\%.
\]
Stratification is performed across:
\begin{itemize}
    \item Steering family (e.g. sine vs. ramp),
    \item Force family (e.g. constant vs. multi-phase),
    \item Speed bucket (e.g. low, medium, high).
\end{itemize}
This ensures that validation and test sets contain balanced dynamic diversity, providing a robust assessment of \gls{pcarnn} generalization to unseen regimes.

\subsection{Geofence Test Scenario Generation and Labeling}
\label{sec:geofence_labeling}

The final test dataset used for closed-loop evaluation is created using a \textbf{guaranteed-solvable random sampling algorithm} designed to generate dynamically valid containment scenarios.

\begin{enumerate}
    \item \textbf{Randomized initialization:}
    The vehicle’s initial position $(p_x, p_y, \psi)$ is uniformly sampled within the geofence region $\Omega$, producing diverse proximity and approach angles relative to the boundary.
    \item \textbf{Solvability check:}
    Each sampled initial state undergoes a feasibility test using a \textit{panic brake baseline}.
    The scenario is retained only if applying maximum braking $F_{x,\min}$ can bring the vehicle to a complete stop within $\Omega$, ensuring physical solvability.
    \item \textbf{Test trajectory labeling:}
    Each valid scenario is executed under nominal dynamics (without \gls{dcbf} intervention) and labeled:
    \begin{itemize}
        \item \textbf{Unsafe:} if the trajectory exits $\Omega$ at any time.
        \item \textbf{Safe:} otherwise.
    \end{itemize}
    \item \textbf{Final state capture:}
    Each run concludes with a forced braking phase until the vehicle comes to rest.
    The complete trajectory, including states, controls, and safety labels, is stored as a Parquet file for computing \gls{cf1} and \gls{fpr} metrics during evaluation.
\end{enumerate}

\section{Breaking Only Geofencing Results}
\begin{table*}[ht]
\centering
\caption{Summary of best-performing variants (highest CF1) from each model group under \textbf{braking-only} settings for both vehicles. One small ($\sim$50K) and one large ($\sim$150K) configuration are selected per variant type. Lower is better for FPR and MCD, higher is better for CF1.}
\resizebox{1\textwidth}{!}{
\begin{tabular}{
  p{1.5cm}  
  p{2.5cm}  
  p{1.8cm}  
  p{4.5cm}  
  p{2.1cm}  
  p{2.1cm}  
  p{2.1cm}  
  p{2.1cm}  
}
\toprule
\textbf{Vehicle} & \textbf{Model Variant} & \textbf{Controller} & \textbf{Size ($f$, $g$)} & \textbf{Params} & \textbf{CF1}${\uparrow}$ & \textbf{FPR}${\downarrow}$ & \textbf{MCD}$^{+}{\downarrow}$ \\
\midrule

\multirow{6}{*}{Audi E-tron}
  & PCARNN (shared) & TTC / TTC & $4\times128$ / $5\times192$ & 51.3K / 150K & {0.800} / {0.802} & 0.164 / 0.175 & 2.575 / 2.690 \\
  & PCARNN  (split)& TTC / TTC & $(3\times90, 4\times104)$ / $(5\times135, 5\times135)$ & 51K / 149K & 0.787 / \underline{0.803} & 0.257 / {0.186} & 4.294 / {1.800} \\
  & Residual & DCBF / DCBF & $4\times128$ / $5\times192$ & 50.9K / 150K & {0.604} / 0.571 & 0.082 / \underline{0.066} & \underline{0.207} / 0.236 \\
  & Neural ODE & TTC / DCBF & $4\times128$ / $5\times192$ & 50.9K / 150K & 0.705 / {0.522} & 0.131 / \textbf{0.060} & -0.912 / \textbf{0.136} \\
  & Bicycle & TTC  & N/A & N/A & \textbf{0.816} & 0.137 & 2.084 \\
  & Bicycle Ackermann & BRT & N/A & N/A & {0.727} & 0.284 & 2.379 \\
\midrule

\multirow{6}{*}{Lincoln MKZ}
  & PCARNN (shared) & TTC / TTC & $4\times128$ / $5\times192$ & 51.3K / 150K & 0.714 / \textbf{0.806} & 0.088 / 0.106 & 0.466 / 0.574 \\
  & PCARNN  (split)& TTC / DCBF & $(3\times90, 4\times104)$ / $(5\times156, 4\times128)$ & 51K / 150K & 0.718 / {0.640} & 0.088 / \textbf{0.050} & 0.466 / {0.411} \\
  & Residual & DCBF / DCBF & $4\times128$ / $5\times192$ & 50.9K / 150K & 0.657 / {0.624} & 0.069 / \underline{0.056} & 0.452 / \underline{0.389} \\
  & Neural ODE & DCBF / TTC & $4\times128$ / $5\times192$ & 50.9K / 150K & 0.657 / {0.651} & 0.069 / 0.106 & 0.452 / \textbf{0.229} \\
  & Bicycle & TTC & N/A & N/A & \underline{0.783} & 0.131 & 0.673 \\
  & Bicycle Ackermann & BRT & N/A & N/A & {0.672} & 0.206 & 0.778 \\

\bottomrule
\end{tabular}}
\end{table*}

\begin{table*}[ht]
\centering
\caption{Comparison of model architecture variants under different controllers (TTC and DCBF) for braking only scenarios. Lower is better for \gls{fpr} and \gls{mcd}, and higher is better for \gls{cf1}.}
\resizebox{1\textwidth}{!}{
\begin{tabular}{
  p{1.8cm}   
  p{3.6cm}   
  p{3.8cm}   
  p{2.5cm}   
  p{1.5cm}   
  p{1.5cm}   
  p{1.5cm}   
  p{1.5cm}   
}
\toprule
\textbf{Vehicle} & \textbf{Variant} & \textbf{Size} & \textbf{Controller} & \textbf{Params} & \textbf{CF1} ${\uparrow}$  & \textbf{FPR} ${\downarrow}$  & \textbf{MCD}$^{+}{\downarrow}$ \\
\midrule

\multirow{24}{*}{Audi E-tron}
& PCARNN (shared) & \(4\times128\) & TTC & 51.3K & 0.800 & 0.164 & 2.575 \\
& PCARNN (shared) & \(5\times192\) & TTC & 150K  & 0.802 & 0.175 & 2.690 \\
\cline{2-8}\addlinespace[2pt]
& PCARNN  (split)& \(f=3\times90,\; g=4\times104\) & TTC & 51K  & 0.787 & 0.257 & 4.294 \\
& PCARNN  (split)& \(f=5\times135,\; g=5\times135\) & TTC & 149K & 0.794 & 0.175 & 1.643 \\
& PCARNN  (split)& \(f=5\times156,\; g=4\times128\) & TTC & 150K & 0.798 & 0.180 & 1.827 \\
& PCARNN  (split)& \(f=4\times128,\; g=5\times156\) & TTC & 150K & 0.803 & 0.186 & 1.800 \\
\cline{2-8}\addlinespace[2pt]

& Residual & \(4\times128\) & TTC & 50.9K & 0.287 & 0.131 & -0.192 \\
& Residual & \(5\times192\) & TTC & 150K  & 0.449 & 0.093 & -0.023 \\
\cline{2-8}\addlinespace[2pt]

& Neural ODE & \(4\times128\) & TTC & 50.9K & 0.705 & 0.131 & 0.403 \\
& Neural ODE & \(5\times192\) & TTC & 150K  & 0.350 & 0.109 & -0.098 \\
\cline{2-8}\addlinespace[2pt]

& Bicycle & N/A & TTC & N/A & 0.816 & 0.137 & 2.084 \\
\cline{2-8}\addlinespace[4pt]

& PCARNN (shared) & \(4\times128\) & DCBF (\(\gamma=0.40\)) & 51.3K & 0.624 & 0.093 & 0.475 \\
& PCARNN (shared) & \(5\times192\) & DCBF (\(\gamma=0.45\)) & 150K  & 0.595 & 0.071 & 0.411 \\
\cline{2-8}\addlinespace[2pt]

& PCARNN  (split)& \(f=3\times90,\; g=4\times104\) & DCBF (\(\gamma=0.45\)) & 51K  & 0.609 & 0.071 & 0.336 \\
& PCARNN  (split)& \(f=5\times135,\; g=5\times135\) & DCBF (\(\gamma=0.45\)) & 149K & 0.603 & 0.066 & 0.308 \\
& PCARNN  (split)& \(f=5\times156,\; g=4\times128\) & DCBF (\(\gamma=0.45\)) & 150K & 0.609 & 0.071 & 0.319 \\
& PCARNN  (split)& \(f=4\times128,\; g=5\times156\) & DCBF (\(\gamma=0.45\)) & 150K & 0.612 & 0.071 & 0.284 \\
\cline{2-8}\addlinespace[2pt]

& Residual & \(4\times128\) & DCBF (\(\gamma=0.45\)) & 50.9K & 0.537 & 0.060 & 0.088 \\
& Residual & \(4\times128\) & DCBF (\(\gamma=0.40\)) & 50.9K & 0.604 & 0.082 & 0.207 \\
& Residual & \(5\times192\) & DCBF (\(\gamma=0.45\)) & 150K  & 0.571 & 0.066 & 0.236 \\
\cline{2-8}\addlinespace[2pt]

& Neural ODE & \(4\times128\) & DCBF (\(\gamma=0.45\)) & 50.9K & 0.528 & 0.060 & 0.113 \\
& Neural ODE & \(4\times128\) & DCBF (\(\gamma=0.40\)) & 50.9K & 0.556 & 0.077 & 0.203 \\
& Neural ODE & \(5\times192\) & DCBF (\(\gamma=0.45\)) & 150K  & 0.522 & 0.060 & 0.136 \\
\cline{2-8}\addlinespace[2pt]

& Bicycle & N/A & DCBF (\(\gamma=0.40\)) & N/A & 0.622 & 0.098 & 0.454 \\
  \cline{2-8}\addlinespace[2pt]
& Bicycle Ackermann & N/A & BRT & N/A & 0.727 & 0.284 & 2.379 \\
\midrule

\multirow{23}{*}{Lincoln MKZ}
& PCARNN (shared) & \(4\times128\) & TTC & 51.3K & 0.714 & 0.088 & 0.466 \\
& PCARNN (shared) & \(5\times192\) & TTC & 150K  & 0.806 & 0.106 & 0.574 \\
\cline{2-8}\addlinespace[2pt]

& PCARNN  (split)& \(f=3\times90,\; g=4\times104\) & TTC & 51K  & 0.718 & 0.081 & 0.564 \\
& PCARNN  (split)& \(f=5\times135,\; g=5\times135\) & TTC & 149K & 0.562 & 0.094 & 0.249 \\
& PCARNN  (split)& \(f=5\times156,\; g=4\times128\) & TTC & 150K & 0.552 & 0.106 & 0.293 \\
& PCARNN  (split)& \(f=4\times128,\; g=5\times156\) & TTC & 150K & 0.536 & 0.081 & 0.155 \\
\cline{2-8}\addlinespace[2pt]

& Residual & \(4\times128\) & TTC & 50.9K & 0.431 & 0.081 & -0.068 \\
& Residual & \(5\times192\) & TTC & 150K  & 0.623 & 0.106 & 0.215 \\
\cline{2-8}\addlinespace[2pt]

& Neural ODE & \(4\times128\) & TTC & 50.9K & 0.546 & 0.125 & 0.113 \\
& Neural ODE & \(5\times192\) & TTC & 150K  & 0.651 & 0.106 & 0.229 \\
\cline{2-8}\addlinespace[2pt]

                 & Bicycle & N/A & TTC & N/A & 0.783 & 0.131 & 0.673 \\
  \cline{2-8}
\addlinespace[4pt]
& PCARNN (shared) & \(4\times128\) & DCBF (\(\gamma=0.40\)) & 51.3K & 0.676 & 0.069 & 0.486 \\
& PCARNN (shared) & \(5\times192\) & DCBF (\(\gamma=0.45\)) & 150K  & 0.634 & 0.056 & 0.435 \\
\cline{2-8}\addlinespace[2pt]

& PCARNN  (split)& \(f=3\times90,\; g=4\times104\) & DCBF (\(\gamma=0.45\)) & 51K  & 0.634 & 0.056 & 0.477 \\
& PCARNN  (split)& \(f=5\times135,\; g=5\times135\) & DCBF (\(\gamma=0.45\)) & 149K & 0.634 & 0.056 & 0.407 \\
& PCARNN  (split)& \(f=5\times156,\; g=4\times128\) & DCBF (\(\gamma=0.45\)) & 150K & 0.640 & 0.050 & 0.411 \\
& PCARNN  (split)& \(f=4\times128,\; g=5\times156\) & DCBF (\(\gamma=0.45\)) & 150K & 0.640 & 0.050 & 0.422 \\
\cline{2-8}\addlinespace[2pt]

& Residual & \(4\times128\) & DCBF (\(\gamma=0.45\)) & 50.9K & 0.598 & 0.056 & 0.345 \\
& Residual & \(4\times128\) & DCBF (\(\gamma=0.40\)) & 50.9K & 0.657 & 0.069 & 0.452 \\
& Residual & \(5\times192\) & DCBF (\(\gamma=0.45\)) & 150K  & 0.624 & 0.056 & 0.389 \\
\cline{2-8}\addlinespace[2pt]

& Neural ODE & \(4\times128\) & DCBF (\(\gamma=0.45\)) & 50.9K & 0.601 & 0.056 & 0.389 \\
& Neural ODE & \(4\times128\) & DCBF (\(\gamma=0.40\)) & 50.9K & 0.657 & 0.069 & 0.483 \\
& Neural ODE & \(5\times192\) & DCBF (\(\gamma=0.45\)) & 150K  & 0.615 & 0.056 & 0.402 \\
\cline{2-8}\addlinespace[2pt]

& Bicycle & N/A & DCBF (\(\gamma=0.40\)) & N/A & 0.705 & 0.069 & 0.499 \\
    \cline{2-8}\addlinespace[2pt]
& Bicycle Ackermann & N/A & BRT & N/A & 0.672 & 0.206 & 0.778 \\

\bottomrule
\end{tabular}}
\end{table*}

\ifCLASSOPTIONcaptionsoff
  \newpage
\fi

\end{document}

%% file: graph_preamble.tex
\usetikzlibrary{arrows.meta,positioning,shapes.geometric,calc,fit,backgrounds}

\definecolor{inputblue}{RGB}{224, 231, 255}
\definecolor{inputborder}{RGB}{55, 48, 163}
\definecolor{modelyellow}{RGB}{254, 249, 195}
\definecolor{modelborder}{RGB}{161, 98, 7}
\definecolor{processpurple}{RGB}{243, 232, 255}
\definecolor{processborder}{RGB}{88, 28, 135}
\definecolor{qpgreen}{RGB}{220, 252, 231}
\definecolor{qpborder}{RGB}{21, 128, 61}
\definecolor{passthroughblue}{RGB}{224, 242, 254}
\definecolor{passthroughborder}{RGB}{2, 132, 199}
\definecolor{interventiongreen}{RGB}{198, 246, 213}
\definecolor{interventionborder}{RGB}{47, 133, 90}
\definecolor{emergencyred}{RGB}{254, 215, 215}
\definecolor{emergencyborder}{RGB}{197, 48, 48}
\definecolor{outputgray}{RGB}{243, 244, 246}
\definecolor{outputborder}{RGB}{75, 85, 99}

\tikzset{
  >=Latex,
  every node/.style={font=\normalsize},
  input/.style={
    draw=inputborder, line width=1.5pt, rounded corners=4pt, 
    fill=inputblue, align=center, text width=35mm, 
    minimum height=18mm, inner sep=3mm
  },
  model/.style={
    draw=modelborder, line width=1.5pt, rounded corners=4pt,
    fill=modelyellow, align=center, text width=40mm,
    minimum height=18mm, inner sep=3mm
  },
  process/.style={
    draw=processborder, line width=1.5pt, rounded corners=4pt,
    fill=processpurple, align=center, text width=52mm,
    minimum height=25mm, inner sep=3mm
  },
  qp/.style={
    draw=qpborder, line width=1.5pt, rounded corners=4pt,
    fill=qpgreen, align=left, text width=75mm,
    minimum height=28mm, inner sep=4mm
  },
  decision/.style={
    draw=outputborder, line width=1.5pt, diamond, 
    aspect=2.5, align=center, fill=white,
    inner sep=2mm
  },
  passthrough/.style={
    draw=passthroughborder, line width=1.5pt, rounded corners=4pt,
    fill=passthroughblue, align=center, text width=40mm,
    minimum height=14mm, inner sep=3mm
  },
  intervention/.style={
    draw=interventionborder, line width=1.5pt, rounded corners=4pt,
    fill=interventiongreen, align=center, text width=40mm,
    minimum height=14mm, inner sep=3mm
  },
  emergency/.style={
    draw=emergencyborder, line width=1.5pt, rounded corners=4pt,
    fill=emergencyred, align=center, text width=40mm,
    minimum height=14mm, inner sep=3mm
  },
  output/.style={
    draw=outputborder, line width=1.5pt, rounded corners=4pt,
    fill=outputgray, align=center, text width=30mm,
    minimum height=12mm, inner sep=3mm
  }
}

\definecolor{driftgreen}{RGB}{34, 197, 94}
\definecolor{driftgreenbg}{RGB}{220, 252, 231}
\definecolor{controlorange}{RGB}{249, 115, 22}
\definecolor{controlorangebg}{RGB}{254, 237, 213}
\definecolor{neutralgray}{RGB}{243, 244, 246}

\tikzset{
  >=Latex,
  every node/.style={font=\normalsize},
  iobox/.style={draw=black,line width=1.5pt,fill=white,align=center,
    minimum width=18mm,minimum height=15mm,inner sep=2mm},
  driftbox/.style={draw=driftgreen,line width=1.5pt,fill=driftgreenbg,align=center,
    minimum width=65mm,minimum height=13mm,inner sep=2.5mm},
  controlbox/.style={draw=controlorange,line width=1.5pt,fill=controlorangebg,align=center,
    minimum width=65mm,minimum height=13mm,inner sep=2.5mm},
  sumcircle/.style={draw,circle,line width=1.5pt,minimum size=12mm,fill=white,inner sep=0pt},
  funcbox/.style={draw=black,line width=1.5pt,fill=white,align=center,
    minimum width=32mm,minimum height=15mm,inner sep=2mm},
  groupbox/.style={draw=black,line width=1.5pt,dashed,rounded corners=3pt,
    inner xsep=4mm,inner ysep=4mm}
}

%% file: piml_table.tex
\begin{table*}[!htbp]
\centering
\caption{Comparison of models combining physics and machine learning (\gls{piml}) for dynamical systems. These models are illustrated primarily in the context of \glspl{ode}, where $t\in [0, T]$ is the time coordinate, $x(t)$ denotes the system's state vector, $\dot{x}(t)$ is its time derivative, $u(t)$ is the control function, and $x_k$ denotes the measurement indexed by $k$. Moreover, we use $r$ to denote the set of parameters in the governing physics model, and $\theta$ denotes the parameters in the neural networks.}
\label{tab:physics_ml_comparison}
\resizebox{\textwidth}{!}{%
\begin{tabular}{p{3cm}p{3cm}p{3cm}p{7cm}p{5.5cm}}
\toprule
\textbf{Category} & \textbf{Domain Knowledge Requirement} & \textbf{Generalizable to Unseen Scenarios} & \textbf{Mathematical Formulation}  & \textbf{Pros and Cons} \\
\midrule
\textbf{\gls{pinn}} \cite{farea_understanding_2024, cuomo_scientific_2022,cai_physics-informed_2021,cai_physics-informed_2021-1, sahli_costabal_physics-informed_2020, misyris_physics-informed_2020, raissi_physics-informed_2019, cho_separable_nodate} Networks trained with physics-based loss terms enforcing ODE/PDE constraints at collocation points sampled throughout the domain. 
& \textbf{High} \newline Requires the explicit physics equations (e.g. the PDE/ODE). 
& \textbf{Low} \newline The trained network learns the solution $x(t)$ for one specific instance (one set of fixed initial/boundary conditions and specific controls). 
& \textbf{System:} $\dot{x}(t) = f(x(t), u(t), t; r)$ (known physics)\newline
   \textbf{NN learns:} $\hat{x}(t;\theta) \approx x(t)$\newline
   \textbf{Inputs:} $t, u(t)$, boundary/initial conditions\newline
   \textbf{Loss (physics):} 
   $\mathcal{L}_{\text{phys}}(\dot{\hat{x}}(t_j;\theta),  f\big(\hat{x}(t_j;\theta), u(t_j), t_j; r\big)\big)$ \newline
   \textbf{Loss (data):} 
   $\mathcal{L}_{\text{data}}(\hat{x}(t_k;\theta), x_k)$ \newline
   $\dot{\hat{x}}(t)$ is computed through auto differentiation.
& \textbf{Suitable for:} Known physics constraints; limited data\newline
  \textbf{Advantages:} Enforces known physics; data-efficient \newline
  \textbf{Limitations:} Require known physics; new training required for unseen scenarios (e.g. new boundary conditions)\\

\midrule
  \textbf{Surrogate Models} \cite{zhao_surrogate_2023} "Black-box" neural network approximates expensive physics simulations.&                                                                          \textbf{Low} \newline 
  Requires a large dataset of input-output pairs. No explicit equations are needed.
  &
  \textbf{Low} \newline Primarily used for interpolation within the training data domain. & 
  \textbf{System:} $\hat{x}(t+\Delta t; \theta)=f_{\theta}(x(t),u(t),t)$\newline\textbf{NN Learns:} $f_{\theta}$ \newline
  \textbf{Inputs:} $x(t),u(t),t$\newline
  \textbf{Loss (data):} $\mathcal{L}_{\text{data}}\big(x_{t+\Delta t},f_{\theta}(x(t),u(t),t)\big)$ 
  &\textbf{Suitable for:}  Replacing expensive simulators; usually discrete (typical setup of simulators)\newline
  \textbf{Advantages:} Fast surrogate approximations, require large training datasets 
  \newline\textbf{Limitations:} Poor generalizability/extrapolation beyond training scenarios; may violate physics \\
  
  \midrule
   \textbf{Differentiable Physics Simulations} \cite{newbury_review_2024, degrave_differentiable_2019}
  Differentiable physics simulators combined with neural networks for parameter, material, or control policy optimization through gradient-based learning. 
  & \textbf{High} \newline
    Requires explicit governing equations or a differentiable physics engine implementing $f$; full model structure must be specified by the user. 
  & \textbf{Medium} \newline
    Low when calibrating parameters to a single experiment, but can be high when learning shared parameters or control policies across many operating conditions, within the range represented in the training data. 
  & \textbf{System:} $\dot{x}(t) = f\big(x(t), u(t), t; r\big)$\newline
    \textbf{NN learns:} parameters $r_{\theta}$ (system identification) and/or a control policy $u_{\theta}(t)$ or $u_{\theta}(x(t), t)$\newline
    \textbf{Inputs:} current state $x(t)$ (and possibly $t$) to the policy network; initial conditions and model $f$ to the simulator\newline
    \textbf{Loss (data):} 
    $\mathcal{L}_{\text{data}}\big(x_k, \hat{x}\big(t_k; p_{\theta}, u_{\theta}\big)\big)$ where $\hat{x}$ is obtained by integrating $f$ with $p_{\theta}$ and $u_{\theta}$ 
  & \textbf{Suitable for:} Parameter calibration, system identification, and optimal control when high-fidelity differentiable simulators are available.\newline
    \textbf{Advantages:} Guarantees that trajectories follow the specified physics; enables end-to-end gradient-based optimization of parameters and control policies; leverages mature numerical solvers.\newline
    \textbf{Limitations:} Requires accurate and differentiable physics models; computationally expensive due to repeated forward and backward simulations; cannot recover missing physics structure beyond $f$.\\
    
  \midrule
    \textbf{Physics-guided Models} \cite{daw_physics-guided_2021, jia_physics-guided_2021, pawar_physics_2021}
  Hybrid models that combine empirical neural networks with physics-based regularization terms and/or physics-derived features so that predictions respect known constraints or low-order mechanistic relationships. 
  & \textbf{Medium} \newline Requires partial physics knowledge (e.g. constraint function $g$, invariants, or a coarse mechanistic model), but not the full ODE/PDE. 
  & \textbf{Medium} \newline Physics constraints improve extrapolation compared to purely data-driven models, but performance is limited to regimes where the encoded physics remains valid. 
  & \textbf{System:} $\hat{y} = f_{\theta}(x)$\newline
    \textbf{NN learns:} $f_{\theta}$ (the empirical mapping) \newline
    \textbf{Inputs:} features $x$\newline
      \textbf{Loss (physics):} $\mathcal{L}_{\text{phys}}\big(g(f_{\theta}(x),x)\big)$\newline 
    \textbf{Loss (data):} $\mathcal{L}_{\text{data}}\big(y, f_{\theta}(x)\big) $
  & \textbf{Suitable for:} Systems with partially known physics or reliable constraints where full governing equations are unavailable or too complex.\newline
    \textbf{Advantages:} Uses physics to regularize the hypothesis space; can reduce data needs; improves interpretability and physical plausibility; flexible to incorporate diverse constraint types.\newline
    \textbf{Limitations:} Requires expert specification of constraints or mechanistic terms; misspecified $g$ can bias learning; still relies on training data coverage and may extrapolate poorly outside the range where the physics assumptions hold.\\

  \midrule
    \textbf{Neural \glspl{ode}} \cite{tong_neural_2025, worsham_guide_2025, chen_neural_2019}
 Neural networks parameterize the right-hand side of an ODE, and a numerical ODE solver is used as part of the computational graph so that continuous-time dynamics are learned directly from data. 
  & \textbf{Low} \newline Does not require explicit physics equations; only observed trajectories are needed, although domain knowledge can be encoded via architecture or regularization. 
  & \textbf{Medium} \newline Generalizes well to new initial conditions and time grids within the state and control regimes seen during training; extrapolation to qualitatively new regimes is limited. 
  & \textbf{System:} $\hat{\dot{x}}(t; \theta)=f_{\theta}(x(t),u(t),t)$\newline
    \textbf{NN learns:} $f_{\theta}$ (the dynamics)\newline
    \textbf{Inputs:} $u(t),t$, boundary/initial conditions\newline
    \textbf{Loss (data):} $\mathcal{L}_{\text{data}}\big(x(t), \hat{x}(t;\theta)\big)$ \newline
    $\hat{{x}}$ is typically obtained by integrating the derivative $\hat{\dot{x}}$ through a numerical ODE solver (RK4, Dormand–Prince, adaptive solvers, etc.)
  & \textbf{Suitable for:} Modeling continuous-time dynamics and irregularly sampled time series; learning latent dynamical systems. \newline
    \textbf{Advantages:} Handles variable-step integration and irregular sampling; parameter sharing over time; can be combined with latent-variable models for complex systems.\newline
    \textbf{Limitations:} Training can be computationally expensive due to repeated ODE solves; sensitivity to stiff dynamics and solver choices; does not automatically enforce physical constraints unless additional structure or penalties are introduced.\\

\midrule
    \textbf{Residual Models} \cite{el_haloui_combining_2025, rackauckas_universal_2021, bock_hybrid_2021,Yin_2021}
 Neural networks augment a known physics model by learning residual corrections to account for model mismatch, unmodeled effects, or empirical adjustments. 
  & \textbf{Medium} \newline
    Requires a baseline physics model $f$ that captures coarse dynamics; residual network is used to correct remaining discrepancies. 
  & \textbf{High} \newline
    Can generalize well across operating conditions where the baseline model remains qualitatively valid, since the main structure is provided by $f$ and the NN only learns corrections. 
  & \textbf{System:}
    $\hat{\dot{x}}(t; \theta) = f\big(x(t), u(t), t; r\big) + \Delta f_{\theta}\big(x(t), u(t), t\big)$\newline
    \textbf{NN learns:} $\Delta f_{\theta}$ (residual term) that models the discrepancy between the baseline physics and observed dynamics\newline
    \textbf{Inputs:} $x(t), u(t), t$, initial/boundary conditions\newline
    \textbf{Loss (data):} 
    $\mathcal{L}_{\text{data}}\big(x_k, \hat{x}(t_k; \theta)\big)$ 
  & \textbf{Suitable for:} Improving forecasts when a trusted but imperfect physics model exists.\newline
    \textbf{Advantages:} Preserves interpretability and structure of the baseline model; focuses learning capacity on systematic errors; often improves extrapolation relative to purely data-driven models.\newline
    \textbf{Limitations:} Performance depends on the quality of the baseline model; residual may absorb structural errors and become hard to interpret; extrapolation fails if the underlying physics model is invalid in new regimes.\\

  \midrule
  \textbf{Operator Learning Models} \cite{raonic_convolutional_2023, li_fourier_2021, lu_learning_2021}
  Networks learn a solution operator that maps problem definitions (for example parameters, forcing terms, and boundary or initial conditions) to full trajectories or fields, rather than predicting state evolution step by step. 
  & \textbf{Medium} \newline
    Requires specifying a family of ODE/PDE problems, including the type of equation and ranges of parameters and conditions used to generate training data, but not necessarily the closed-form operator. 
  & \textbf{High} \newline
    Once trained on a distribution of parameters and conditions, the learned operator can generalize to unseen configurations within that distribution, enabling rapid evaluation for new scenarios. 
  & \textbf{Formulation:} $x(t) = \mathcal{G}\big(u(\cdot), r, t\big)$, \newline
    \textbf{NN learns:} 
    $\hat{x}(t; \theta) = G_{\theta}\big(u(\cdot), r, t\big)$\newline
    \textbf{Inputs:} discretized or encoded forcing/control functions $u(\cdot)$, problem parameters $r$, initial/boundary conditions, and query times $t$\newline
    \textbf{Loss (data):} 
    $\mathcal{L}_{\text{data}}\big(x_k, \hat{x}(t_k; \theta)\big)$ (often integrated over space and time for PDEs) 
  & \textbf{Suitable for:} Fast emulation of families of physics problems, rapid evaluation under varying parameters or conditions, and surrogate modeling when many training solves are available offline.\newline
    \textbf{Advantages:} Learns global mappings from problem setup to solution; amortizes the cost of expensive solvers over many queries; well suited for real-time or many-query scenarios.\newline
    \textbf{Limitations:} Requires large and diverse training datasets that cover the parameter and condition space; extrapolation outside the training distribution is unreliable; still depends on the correctness of the simulator or data source used for training.\\

\midrule
  \textbf{Data-driven Discovery} \cite{brunton_discovering_2016}
   Methods that infer explicit governing equations directly from data, often via sparse regression or symbolic neural approaches, to recover interpretable dynamical models. 
  & \textbf{High} \newline
    Requires designing a candidate library of basis functions (for example polynomials, nonlinear interactions, control terms) and appropriate sparsity or regularization assumptions; physics structure is then inferred from data. 
  & \textbf{Medium} \newline
    Discovered equations can generalize within the regime represented in the data, but reliability depends strongly on data quality, noise levels, and whether the true dynamics lie in the chosen function library. 
  & \textbf{Formulation:} 
    $\hat{\dot{x}}(t) = \Theta\big(x(t), u(t), t\big)\,\xi$\newline
    where $\Theta(\cdot)$ is a library of candidate nonlinear terms evaluated from data and $\xi$ is a coefficient vector.\newline
    \textbf{NN / optimizer learns:} a sparse coefficient vector $\xi$ (and possibly the structure of $\Theta$) that selects which terms form the governing equations.\newline
    \textbf{Inputs:} measured trajectories $x(t)$, controls $u(t)$, and estimated derivatives $\dot{x}(t)$ (or smoothed approximations)\newline
    \textbf{User sets:} the candidate function library $\Theta$, sparsity-promoting penalties, and selection thresholds\newline
    \textbf{Loss (data + sparsity):} 
    $\mathcal{L}\big(\dot{x}, \Theta(x, u, t)\,\xi\big) + \lambda\|\xi\|_1$ (or related sparse model-selection objectives) 
  & \textbf{Suitable for:} Hypothesis generation, discovering compact analytic models from experimental or simulation data, and gaining mechanistic insight into previously unknown dynamics.\newline
    \textbf{Advantages:} Produces interpretable equations rather than only black-box predictors; can reveal dominant physical mechanisms; useful when explicit models are unavailable or uncertain.\newline
    \textbf{Limitations:} Sensitive to noise, differentiation errors, and the choice of candidate library; often requires dense, high-quality data; less useful when governing equations are already well established and trusted.\\

\bottomrule
\end{tabular}}
\end{table*}

%% file: graph_system.tex
\begin{tikzpicture}[node distance=20mm and 25mm]

\node[input] (x) {
  \large $\boldsymbol{x_k}$\\[2pt]
  \large Current State\\[-1pt]
  \normalsize 
};

\node[input, left=30mm of x] (unom) {
  \large $\boldsymbol{u_{\mathrm{nom}}}$\\[2pt]
  \large Nominal Control
};

\node[input, right=30mm of x] (ulims) {
  \large $\boldsymbol{u_{\min}, u_{\max}}$\\[2pt]
  \large Dynamic Limits\\[-1pt]
  \normalsize 
};

\node[model, below right=18mm and -10mm of x] (model) {
  \large System Model\\[2pt]
  $\dot{x} = \hat{f}(x) + \hat{g}(x)\,u$\\[1pt]
  \normalsize (approximate dynamics)
};

\node[process, below=45mm of x] (predict) {
  \textbf{Step 1: Predict Safety Constraint}\\
  \large $(\mathbf{A}, \mathbf{b})$\\[3pt]
  \large Simulate future trajectories using the model to find the safest intervention boundary
};

\node[qp, below=15mm of predict] (qp) {
  \textbf{Step 2: Solve Safety QP}\\[4pt]
  \large $\displaystyle\min_{\mathbf{u},\,\boldsymbol{\xi} \geq 0} \,
   \tfrac{1}{2}\|\mathbf{u} - \mathbf{u}^{\mathrm{nom}}\|_W^2 + 
   \tfrac{1}{2}\rho\|\boldsymbol{\xi}\|_2^2$\\[4pt]
  $\mathrm{s.t.} \quad 
   \mathbf{A}\mathbf{u} + \boldsymbol{\xi} \geq \mathbf{b}, \quad
   \mathbf{u}_{\min} \leq \mathbf{u} \leq \mathbf{u}_{\max}$
};

\node[decision, below=15mm of qp] (mux) {
  \textbf{Step 3: Intervention}\\
  \textbf{Needed?}
};

\node[passthrough, below left=20mm and 35mm of mux] (nochange) {
  \large $u_{\mathrm{final}} = u_{\mathrm{nom}}$\\[1pt]
  \normalsize (No intervention)
};

\node[intervention, below=20mm of mux] (usafe) {
  \large $u_{\mathrm{final}} = u_{\mathrm{safe}}$\\[1pt]
  \normalsize (Minimal intervention)
};

\node[emergency, below right=20mm and 35mm of mux] (emerg) {
  \large $u_{\mathrm{final}} = [0, u_{\min,F_x}]$\\[1pt]
  \normalsize (Emergency brake)
};

\node[output, below=45mm of mux] (output) {
  Apply to CARLA
};

\draw[->, line width=1pt] (x.south) 
  to[out=-90, in=90] (model.north);

\draw[->, line width=1pt] (x.south) -- (predict.north);

\draw[->, line width=1pt] (unom.south) 
  to[out=-90, in=150] (predict.west);

\draw[->, line width=1pt] (model.south) 
  to[out=-90, in=30] 
  node[right, pos=0.4] {\normalsize Used by} 
  (predict.north east);

\draw[->, line width=1pt] (predict) -- (qp);

\draw[->, line width=1pt] (unom.south) 
  to[out=-90, in=135] 
  node[left, pos=0.3] {\normalsize Target} 
  ([xshift=-25mm]qp.north);

\draw[->, line width=1pt] (ulims.south) 
  to[out=-90, in=45] 
  node[right, pos=0.3] {\normalsize Constraints} 
  ([xshift=25mm]qp.north);

\draw[->, line width=1pt] (qp) -- (mux);

\draw[->, line width=1pt] (mux.south west) 
  to[out=-135, in=90]
  node[above left, pos=0.25, inner sep=1pt, align=center] {
    \normalsize QP feasible \&\\
    \normalsize $u_{\mathrm{safe}} \approx u_{\mathrm{nom}}$
  } (nochange.north);

\draw[->, line width=1pt] (mux) -- 
  node[left, pos=0.4, align=right] {
    \normalsize QP feasible \&\\
    \normalsize $u_{\mathrm{safe}} \neq u_{\mathrm{nom}}$
  } (usafe);

\draw[->, line width=1pt] (mux.south east) 
  to[out=-45, in=90]
  node[above right, pos=0.25, inner sep=1pt] {
    \normalsize QP infeasible
  } (emerg.north);

\draw[->, line width=1pt] (nochange.south) 
  to[out=-90, in=180] (output.west);

\draw[->, line width=1pt] (usafe) -- (output);

\draw[->, line width=1pt] (emerg.south) 
  to[out=-90, in=0] (output.east);

\end{tikzpicture}

%% file: graph_ca_model_split.tex
\begin{tikzpicture}[x=1mm,y=1mm]

\def\xX{-25}
\def\xA{55}
\def\xS{130}
\def\xH{170}
\def\xO{235}

\def\yC{8}

\def\yF{45}
\def\yG{25}
\def\yFN{-12}
\def\yGN{-32}
\def\ySF{20}
\def\ySG{-8}
\def\yU{-32}   

\node[iobox] (x) at (\xX,\yC) {\Large\bfseries x};

\node[driftbox]   (fdrift)   at (\xA,\yF) {Drift dynamics $f_{\text{phys}}(\mathbf{x})$};
\node[controlbox] (gcontrol) at (\xA,\yG) {Control influence $g_{\text{phys}}(\mathbf{x})$};
\begin{scope}[on background layer]
\node[groupbox, fit=(fdrift)(gcontrol),
      label={[align=center]above:\textbf{Dynamic bicycle model}}] (analytical) {};
\end{scope}

\node[driftbox]   (nndrift)   at (\xA,\yFN) {Drift residual $\Delta_{f_{NN}}(\mathbf{x})$};
\node[controlbox] (nncontrol) at (\xA,\yGN) {Control residual $\Delta_{g_{NN}}(\mathbf{x})$};
\begin{scope}[on background layer]
\node[groupbox, fit=(nndrift)(nncontrol),
      label={[align=center]above:\textbf{Data-driven neural network}}] (neural) {};
\end{scope}

\node[sumcircle] (sumf) at (\xS,\ySF) {\Large $+$};
\node[sumcircle] (sumg) at (\xS,\ySG) {\Large $+$};
\node[funcbox] (fhat) at (\xH,\ySF) {\large $\hat{f}(\mathbf{x})$};
\node[funcbox] (ghat) at (\xH,\ySG) {\large $\hat{g}(\mathbf{x})$};

\node[iobox] (u) at (\xH,\yU) {\Large\bfseries u};

\node[funcbox, minimum width=60mm, minimum height=18mm]
  (output) at (\xO,\yC) {$\dot{\mathbf{x}} = \hat{f}(\mathbf{x}) + \hat{g}(\mathbf{x})\,\mathbf{u}$};

\draw[->, line width=1.2pt] (x.east) to[out=0, in=180] (fdrift.west);
\draw[->, line width=1.2pt] (x.east) to[out=0, in=180] (gcontrol.west);
\draw[->, line width=1.2pt] (x.east) to[out=0, in=180] (nndrift.west);
\draw[->, line width=1.2pt] (x.east) to[out=0, in=180] (nncontrol.west);

\draw[->, line width=1.5pt, color=driftgreen]    (fdrift.east)   to[out=0, in=135]  (sumf.north west);
\draw[->, line width=1.5pt, color=controlorange] (gcontrol.east) to[out=0, in=135]  (sumg.north west);
\draw[->, line width=1.5pt, color=driftgreen]    (nndrift.east)   to[out=0, in=-135] (sumf.south west);
\draw[->, line width=1.5pt, color=controlorange] (nncontrol.east) to[out=0, in=-135] (sumg.south west);
\draw[->, line width=1.5pt, color=driftgreen]    (sumf.east) -- (fhat.west);
\draw[->, line width=1.5pt, color=controlorange] (sumg.east) -- (ghat.west);
\draw[->, line width=1.5pt, color=driftgreen]    (fhat.east) to[out=0, in=90]  (output.north);
\draw[->, line width=1.5pt, color=controlorange] (ghat.east) to[out=0, in=-90] (output.south);
\draw[->, line width=1.5pt, color=controlorange] (u.north) to[out=90, in=-90] (ghat.south);

\end{tikzpicture}

%% file: graph_ca_model_shared.tex
\begin{tikzpicture}[x=1mm,y=1mm]

\def\xX{-25}
\def\xA{55}
\def\xS{130}
\def\xH{170}
\def\xO{235}

\def\yC{8}

\def\yF{45}
\def\yG{25}
\def\yFN{-12}
\def\yGN{-32}
\def\ySF{20}
\def\ySG{-8}
\def\yU{-32}

\node[iobox] (x) at (\xX,\yC) {\Large\bfseries x};

\node[driftbox]   (fdrift)   at (\xA,\yF) {Drift dynamics $f_{\text{phys}}(\mathbf{x})$};
\node[controlbox] (gcontrol) at (\xA,\yG) {Control influence $g_{\text{phys}}(\mathbf{x})$};
\begin{scope}[on background layer]
\node[groupbox, fit=(fdrift)(gcontrol),
      label={[align=center]above:\textbf{Dynamic bicycle model}}] (analytical) {};
\end{scope}

\def\yMODEL{-22}

\node[draw=black, line width=1.5pt, fill=white, align=center,
      minimum width=65mm, minimum height=33mm, inner sep=2.5mm] (model) at (\xA,\yMODEL) {
  \large \textbf{Data-driven neural network}\\[3pt]
  $\mathcal{N}_\theta: \mathbf{x}\rightarrow\begin{bmatrix}\Delta_{\theta}{f_{NN}}(\mathbf{x})\\[2pt]\Delta_{\theta}{g_{NN}}(\mathbf{x})\end{bmatrix}$
};

\begin{scope}[on background layer]
\node[groupbox, fit=(model), inner xsep=4mm, inner ysep=4mm,
      label={[align=center]above:\textbf{Data-driven neural network}}] (neural) {};
\end{scope}

\node[sumcircle] (sumf) at (\xS,\ySF) {\Large $+$};
\node[sumcircle] (sumg) at (\xS,\ySG) {\Large $+$};
\node[funcbox] (fhat) at (\xH,\ySF) {\large $\hat{f}(\mathbf{x})$};
\node[funcbox] (ghat) at (\xH,\ySG) {\large $\hat{g}(\mathbf{x})$};

\node[iobox] (u) at (\xH,\yU) {\Large\bfseries u};

\node[funcbox, minimum width=60mm, minimum height=18mm]
  (output) at (\xO,\yC) {$\dot{\mathbf{x}} = \hat{f}(\mathbf{x}) + \hat{g}(\mathbf{x})\,\mathbf{u}$};

\draw[->, line width=1.2pt] (x.east) to[out=0, in=180] (fdrift.west);
\draw[->, line width=1.2pt] (x.east) to[out=0, in=180] (gcontrol.west);

\draw[->, line width=1.2pt] (x.east) to[out=0, in=180] (model.west);

\draw[->, line width=1.5pt, color=driftgreen]    (fdrift.east)   to[out=0, in=135]  (sumf.north west);
\draw[->, line width=1.5pt, color=controlorange] (gcontrol.east) to[out=0, in=135]  (sumg.north west);

\path let \p1 = (model.east) in
      coordinate (modelEup) at (\x1,\y1+2)
      coordinate (modelEdn) at (\x1,\y1-2);
\draw[->, line width=1.5pt, color=driftgreen]
  (modelEup) to[out=0, in=-135] (sumf.south west);
\draw[->, line width=1.5pt, color=controlorange]
  (modelEdn) to[out=0, in=-135] (sumg.south west);

\draw[->, line width=1.5pt, color=driftgreen]    (sumf.east) -- (fhat.west);
\draw[->, line width=1.5pt, color=controlorange] (sumg.east) -- (ghat.west);

\draw[->, line width=1.5pt, color=driftgreen]    (fhat.east) to[out=0, in=90]  (output.north);
\draw[->, line width=1.5pt, color=controlorange] (ghat.east) to[out=0, in=-90] (output.south);

\draw[->, line width=1.5pt, color=controlorange] (u.north) to[out=90, in=-90] (ghat.south);

\end{tikzpicture}